\documentclass[journal,peerreview,nodraftcls]{IEEEtran}
\usepackage{graphicx,subfig,amsthm,amsmath,latexsym,amssymb,times,bm}
\usepackage{float,epsfig,multirow,rotating,times,verbatim,wrapfig,booktabs}
\usepackage{color,xr,array,hyperref,chngcntr}

  \usepackage{cite}

\usepackage{amsmath}

\DeclareMathOperator{\arcsinh}{arcsinh}

\newtheorem{thm}{Theorem}

\newtheorem{result}[thm]{Result}

\newcommand{\R}{\mathbb{R}}

\newcommand{\mW}{\mathcal W}

\newcommand{\mC}{\mathcal C}
\newcommand{\mbC}{\boldsymbol{\mC}}
\newcommand{\mK}{\mathcal K}

\newcommand{\mL}{\mathcal L}

\newcommand{\E}{\mathbb{E}}

\newcommand{\abs}[1]{|#1|}
\newcommand{\norm}[1]{\Vert#1\Vert}

\newcommand{\bxi}{\boldsymbol{\xi}}

\newcommand{\bpsi}{\boldsymbol{\psi}}

\newcommand{\bmu}{\boldsymbol{\mu}}

\newcommand{\bX}{\boldsymbol{X}}

\newcommand{\bY}{\boldsymbol{Y}}

\newcommand{\bmX}{\boldsymbol{\mathcal X}}

\newcommand{\argmax}{\operatornamewithlimits{argmax}}
\newcommand{\argmin}{\operatornamewithlimits{argmin}}

\newcommand{\A}{\boldsymbol{A}}
\newcommand{\1}{\boldsymbol{1}}

\newcommand{\C}{\mathcal{C}}
\renewcommand{\L}{\mathcal{L}}
\newcommand{\m}{\boldsymbol{\mu}}
\newcommand{\X}{\boldsymbol{X}}
\newcommand{\Y}{\boldsymbol{Y}}

\newcommand{\bit}{\begin{itemize}}
\newcommand{\eit}{\end{itemize}}
\newcommand{\ben}{\begin{enumerate}}
\newcommand{\een}{\end{enumerate}}
\newcommand{\beq}{\begin{equation}}
\newcommand{\eeq}{\end{equation}}
\newcommand{\bea}{\begin{eqnarray*}}
\newcommand{\eea}{\end{eqnarray*}}
\newcommand{\bpf}{\begin{proof}}
\newcommand{\epf}{\end{proof}\ms}
\newcommand{\ms}{\medskip}

\newcommand{\citep}{\cite}
\newcommand{\citet}{\cite}
\newcommand{\citeyear}{\cite}
\newcommand{\citeauthor}{\cite}

\begin{document}

\title{An efficient $k$-means-type  algorithm for clustering datasets with incomplete records}%
\author{Andrew~Lithio~and~Ranjan~Maitra
  \IEEEcompsocitemizethanks{\IEEEcompsocthanksitem A. Lithio is with 
    Statistics--Oncology, Eli Lilly and Company, Indianapolis, Indiana
    46285, USA. Email: lithio\_andrew@lilly.com. 
\IEEEcompsocthanksitem R. Maitra is with
    the Department of Statistics, Iowa State University, Ames, Iowa 50011-1090, USA. Email: maitra@iastate.edu.}
    \thanks{R. Maitra's research was supported in part by the
    National Institute of 
Biomedical Imaging and Bioengineering (NIBIB) of the National
Institutes of Health (NIH) under its Award No. R21EB016212,
The content of this paper however is solely the responsibility of the 
authors and does not represent the official views of either the
NIBIB or the NIH.}

}
\markboth{
}%
{Lithio and Maitra \MakeLowercase{\textit{et al.}}: Efficient $k_m$-means for missing data problems}


\IEEEcompsoctitleabstractindextext{%

\begin{abstract}
The $k$-means algorithm is arguably the most popular nonparametric clustering
method but cannot generally be applied to datasets with incomplete
records. The usual practice then is to either impute missing
values under an assumed missing-completely-at-random
mechanism or to ignore the incomplete records, and apply the
algorithm on the resulting dataset. We develop an efficient version of
the $k$-means algorithm that allows for clustering in the presence of
incomplete records. Our extension is called $k_m$-means and reduces 
to the $k$-means algorithm when all records are complete. We also provide
initialization strategies for our algorithm and methods to estimate
the number of groups in the dataset. Illustrations and simulations
demonstrate the  efficacy of our approach in a variety of settings and
patterns of missing data. Our methods are also applied to  
the analysis of activation images obtained from a functional Magnetic
Resonance Imaging experiment.    

\end{abstract}

\begin{IEEEkeywords}
\sc{Amelia}, {\sc CARP}, fMRI, imputation, jump statistic, 
$k$-means++, $k$-POD, {\sc mice}, soft constraints, SDSS
\end{IEEEkeywords}}
\maketitle

\IEEEdisplaynotcompsoctitleabstractindextext

\section{Introduction}
The need for partitioning or clustering datasets arises in many
diverse applications  
\citep{michenerandsokal57, hinneburgandkeim99, maitra01, feigelsonandbabu98} and has a
long history \citep{jain2010data, hartigan1975clustering, macqueen1967some, lloyd1982least,
kettenring06, xuandwunsch09, melnykovandmaitra10,
maitraetal12}. There is substantial literature on the topic, with
development on the  computational challenges \citep{zhao2009parallel}
as well as on data-driven extensions such as semi-supervised
clustering \citep{basu2004probabilistic} 
and dimension reduction \citep{roweis2000nonlinear}. Datasets often
have missing values   in some features, or variables, presenting
another obstacle for common clustering algorithms and software
packages. Two convenient approaches in such situations are  {\em
  marginalization} and {\em imputation} 
\citep{wagstaff2005making, little2014statistical},  
both of which permit the use of traditional clustering algorithms
without further modification. Marginalization or {\em deletion}
removes any observation record  missing a value in at least one
feature. An alternative approach, used by some
authors~\citep{chattopadhyay2017gaussian}, removes  entirely from
clustering  consideration features that are unobserved for any case. Hybrid
methods of these two deletion schemes  also exist. Yet another {\em
  whole-data strategy}  
\citep{hathawayandbezdek01,chattopadhyayandmaitra18} clusters the
complete records and classifies the incomplete records with rules
based on the obtained grouping and a partial distance or marginal
posterior probability  approach. This scheme inherently assumes
a missing-completely-at-random  (MCAR) mechanism for the unobserved
records and features.  
On the other hand, imputation \citep{honaker2011amelia,
  buuren2011mice, donders2006review, parketal16} predicts the missing
values, and then  assumes that those predictions are as good as
observations and  indistinguishable from the observed data. Because the imputed
observations are treated no differently from the complete records, the
assumptions used in  imputing the values are of critical
importance. Indeed, \citet{wagstaff2005making} illustrate how
imputation can substantially degrade performance when model
assumptions are violated.  

A third approach groups partially unobserved data by developing methods
that inherently incorporate the incompletely-obsereved structure of the
data. Such methods use soft constraints 
\citep{sarkarandleong01,wagstaff2004clustering}, rough sets
\citep{siminski2013clustering} or {\em partial distance} \citep{dixon79}
that is employed by the $k$-means algorithm of
\citet{himmelspachandconrad10} or in the classification  
step of \citet{hathawayandbezdek01}. \citet{sarkarandleong01} modify
fuzzy clustering by  
estimating distances between cluster prototypes and incomplete
observations. Another approach to incorporating missing values in fuzzy clustering \citep{siminski2013clustering, 
simnski2014rough, siminski2015rough} estimates the cluster centers from the completely observed
records and then imputes multiple values for each missing
observation. Lower weights are assigned  
to the augmented observations, which are then included in the
objective function. The $k$-means algorithm with soft constraints
(KSC) of  \citet{wagstaff2004clustering} also separates the   
datasets into two sets, one of the completely observed  and the other
of the partially observed  
features, respectively. The partially observed features are used to
create soft  
constraints that are added to the objective function, essentially acting as an additional 
penalty. This penalty depends on a user-specified weight that \citet{wagstaff2005making} 
suggests should be determined using {\em a priori} knowledge on the importance of the 
partially observed features, or tuned using a labeled subset of data. This methodology works 
only when all records have complete information on at least one feature. \citet{himmelspachandconrad10}
analyze performance of several fuzzy and $k$-means clustering algorithms on two synthetic 
datasets and show that a $k$-means approach using partial distance is the best performer.
Most recently, \citet{chi2016k} developed a majorization-minimization
\citep{hunterandlange04, 
lange16} approach called $k$-POD that can essentially be understood as an iterative imputation
approach, with the current cluster means as the (current) imputed values. Each
iteration clusters the augmented data using $k$-means and then updates
the imputed values through 
the cluster means. The $k$-POD algorithm is implemented in the
R~\citep{R} package  {\sc kpodcluster} \citep{kpodclustr} and
initialized using the $k$-means++ algorithm~\citep{arthur2007k} on the
dataset with missing values imputed from the (global) feature means.  At termination, $k$-POD locally minimizes the objective
function of the $k$-means algorithm using partial distances. However, the repeated application
of $k$-means at every iteration is computationally expensive. The
literature is also sparse on estimating the   number of groups $K$ for data with
incomplete records. 

This paper develops an efficient $k$-means-type clustering
algorithm called {\em $k_m$-means} that  accommodates incomplete
records and generalizes the algorithm of
\citet{hartigan1979algorithm} that is popular in the statistical
literature and software. Expressions for the objective function and
its changes following the cluster reassignment  of an observation play
central roles in our generalization of the
\citet{hartigan1979algorithm}  algorithm. Section~\ref{methods}  also
provides an initialization strategy for $k_m$-means and an  adaptation of
the jump statistic \citep{sugar2003finding} for  estimating the number
of groups. 
Section~\ref{sim} comprehensively evaluates our methodology  through a
series of large-scale simulation   
experiments for datasets of different clustering complexities, sizes,
numbers of groups, and with different missingness mechanisms and
proportions.  Section~\ref{application} uses our methods
to 
find the types of activated cerebral regions from several single-task
functional  Magnetic Resonance Imaging (fMRI) experiments. We conclude
with some  discussion in Section~\ref{discussion}. This paper also has
an online 
supplement having  additional illustrations on performance evaluations
and other preliminary  data analysis. Figures in the supplement
referred to in this paper have the prefix ``S-''.
\section{Methodology}\label{methods}
\subsection{Preliminaries}\label{prelim}
Let $\bmX = \{\X_1,\X_2,\ldots,\X_n\}$ be observation records of $p$
features with  each $\X_i$ possibly missing recorded values for some features. Let $\Y_i$ be a binary vector of $p$ coordinates
with $j$th element $Y_{ij} = I(X_{ij}\mbox{ is recorded})$, where $X_{ij}$ is the $j$th element
of $\X_i$ and $I(\cdot)$ is the indicator function taking value 1 if the function argument is
true and 0 otherwise.  Let $p_i = \sum_{j=1}^p Y_{ij}$ be the number of recorded
features for $\X_i$. For now, we assume that $K$ is known. Our objective is to find the
partition $\mbC = \{ \C_1,\C_2,\ldots,\C_K\}$ with centers $\m_1,\m_2,\ldots,\m_K$  that minimizes 
\begin{equation}\label{wss}
  \mW_K = \sum_{k=1}^K \sum_{i=1}^n \sum_{j=1}^p I(\X_i \in \C_k) Y_{ij} (X_{ij}-\mu_{kj})^2.
\end{equation}
For any given partition $\mbC$, \eqref{wss} is minimized at $$\hat{\mu}_{kj} =  
\frac{\sum_{i=1}^n I(\X_i \in \C_k)Y_{ij}X_{ij}}{\sum_{i=1}^n I(\X_i
  \in \C_k)Y_{ij}}.$$ For completely observed datasets, $Y_{ij}=1$ $\forall$ $i,j$, and $\mW_K$ is the usual 
within-cluster sum of squares (WSS). With incomplete records, the use of $\mW_K$ as the objective
function can be motivated using homogeneous spherical Gaussian and nonparametric
distributional assumptions, as we show next.
\begin{result}\label{result1}
  Suppose that we have Gaussian-distributed observations with
  homogeneous spherical dispersions.
  That is, given $\bX_i\in \C_k$, suppose that each   $X_{ij} \overset{ind}{\sim} N(\mu_{kj} ,
  \sigma^2)$. Then, given the correct partitioning, minimizing \eqref{wss} is equivalent to
  maximizing the loglikelihood function of the parameters $\bmu_1,\bmu_2,\ldots,\bmu_K$ and
  $\sigma$ given the observed $X_{ij}$s. For $j=1,2,\ldots,p$ and
  $k=1,2,\ldots,K$, this optimal value is attained at  
  \begin{equation}\label{mles}
    \hat\mu_{kj} =  \dfrac{\sum_{i=1}^n I(\X_i \in \C_k)Y_{ij}X_{ij}}{\sum_{i=1}^n I(\X_i \in
      \C_k)Y_{ij}} \mbox{ and } \hat\sigma^2 = \mW_K/{\sum_{i=1}^n p_i}.
  \end{equation}
\end{result}
\begin{proof}
  The  loglikelihood function of $(\sigma, \mbC,
 \bmu_1,\ldots,\bmu_K)$ 
is, but for an additive
 constant not   dependent on those parameters, given by $\ell
 (\m_1,\m2,\ldots,\m_K,\sigma^2,\mbC | 
  \bmX) =  -\sum_{i=1}^n  {p_i \log \sigma}/{2} - \sum_{k=1}^K
  \sum_{i=1}^n \sum_{j=1}^p {Y_{ij}I(\X_i \in \C_k) (X_{ij} - \mu_{kj})^2}/{(2
  \sigma^2)}.$ For a given $\mbC$, the second term is free of
  $\bmu_1,\bmu_2,\ldots,\bmu_K$ and $\sigma$ at the maximizing likelihood estimates given by
  \eqref{mles} \citep{johnsonandwichern13}. Then finding the partition  minimizing $\hat{\sigma}$ is
  equivalent to maximizing the profile loglikelihood over all $\bmu_1,\bmu_2,\ldots,\bmu_K$, $\sigma$, 
  and $\mbC$.
\end{proof}
The $k$-means algorithm does not make distributional
assumptions but may be cast in a semi-parametric framework
\citep{maitraandramler09,maitraetal12}. We now show that the following holds
even  without the Gaussian distributional assumptions that underlie Result~\ref{result1}:
\begin{result}\label{result2}
  Suppose that given $\bY_1,\bY_2,\ldots,\bY_n$ and the true partitioning $\mbC$, the first
  two conditional central moments of each $X_{ij}$ are free of $Y_{ij}$ and the $k$ for
  which $\bX_i\in\C_k$. That is, let $\E[(X_{ij}-\mu_{kj})^2\mid\bX_i\in\C_k,Y_{ij}] =
  \sigma^2$, where $\E[X_{ij}\mid\bX_i\in\C_k,Y_{ij}] = \mu_{kj}$. Then $\E[\mW_k\mid
  \bY_1,\bY_2,\ldots,\bY_n] = (n \bar{p} - Kp) \sigma^2$, where $\bar p = \sum_{i=1}^np_i/n$. Thus,
  minimizing $\mW_k$, after conditioning on   $\bY_1,\bY_2,\ldots,\bY_K$ and the true clustering
  $\mbC$, is equivalent in expectation  to minimizing an unbiased estimator for $\sigma^2$. 
\end{result}
\begin{proof}
  Let the number of observations assigned to cluster $k$ be $$n_{kj} =
  \sum_{i=1}^n I(\X_i \in \C_k)Y_{ij}.$$ We assume that $n_{kj} \ge 1$ for every combination of $k$ and $j$. 
  From the assumptions in the theorem, we have 
  \begin{equation*}
    \begin{split}
      \E[\mW_k & \mid \bY_1,\bY_2,\ldots,\bY_n] \\
      &= \sum_{j=1}^p \sum_{k=1}^K \sum_{i=1}^n
      Y_{ij}\E[I(\X_i \in \C_k)  (X_{ij}-\hat{\mu}_{kj})^2 \mid  Y_{ij}  ]\\
      &= \sum_{j=1}^p \sum_{k=1}^K (n_{kj}-1) \sigma^2
      = (n \bar{p} - Kp) \sigma^2.  
    \end{split}
  \end{equation*}
  A similar result with minor modifications holds if some $n_{kj}=0$.
\end{proof}

We now make a few comments in light of Result~\ref{result2}.
\ben
\item Result~\ref{result2} shows that as long as each feature in each group has the same
  conditional variance $\sigma^2$, the missingness mechanism does not, on the average, impact
  the objective function \eqref{wss}. This is a stronger statement than
  Result~\ref{result1}.

\item \citet{timm1998fuzzy}  contend that minimizing ~\eqref{wss} can lead to bias
  in fuzzy clustering. Therefore, they add a ``correction term'' to replace the
  missing features with the value of the corresponding cluster center
  plus an error term in order to try to more accurately represent the
  distance between the cluster center and the complete 
  record. In our view, adding such a term for $k$-means clustering is
  unnecessary because it optimizes the same 
  objective function in expectation as \eqref{wss} and including
  the pseudo-random realizations would add uncertainty   in the
  computations, potentially impeding the algorithm's convergence and
  stability. 
  
\item The objective function of \citet{chi2016k} is also effectively
 $\mW_K$. For let $\X$ be the $n \times p$ matrix of
  observed data, $\m$ be the $k \times p$ matrix of  
  cluster centers, and $\A$ be an $n \times k$ matrix indicating
  cluster membership of each 
  observation. We write that $\A$ is a member of the set $H = \lbrace \A \in \{ 0,1 \}^{n
    \times k} : \A \1 = \1 \rbrace$. Then the objective function for completely observed data is
  $\min_{\A \in H, \m} || \X - \A \bmu ||^2_F,$ where $||\X||_F^2 = \sum_{i,j} x_{ij}^2$
  denotes the Frobenius norm. For partially observed data, let $\Omega = \{(i,j) : Y_{ij} = 1
  \}$ and define the projection operator of any $n \times p$ matrix
  $\bX$ onto $\Omega$ as
  $[P_{\Omega} (\X)]_{ij} = Y_{ij}X_{ij}.$ Then \citet{chi2016k} argue that $\min_{\A \in H,
    \m} || P_{\Omega} (\X) - P_{\Omega} (\A \bmu) ||^2_F = \mW_K$ is
  the natural objective   function for partially observed data. 

\item Operationally, the approach of  \citet{chi2016k} is the same as replacing $X_{ij}$ in
  $\mW_K$ with $\hat{\mu}_{kj}$ for any $Y_{ij}=0$, and then using the $k$-means algorithm at
  every iteration.

  \een 
  
\citet{chi2016k}'s   use of $k$-means at every iteration is
computationally demanding,   so we develop an algorithm that
eliminates the need for such iterations   within an iteration  and
also reduces required computations  only to recently-updated groups
and observations.  

  \subsection{A Hartigan-Wong-type algorithm for clustering with incomplete records}
  \label{hw}
  The \citet{hartigan1979algorithm} $k$-means  clustering algorithm for data with no missing values relies on the  quantities  $\Delta_{k,i}^{\bullet -}$ and $\Delta_{l,i}^{\bullet  +}$, which are, respectively,  the decrease in WSS from removing  $\bX_i$ from cluster $\C_k$, and the increase in   WSS upon adding  observation $\bX_i$ to cluster $\C_l$.
Then  $\Delta_{k,i}^{\bullet -} = {n^\bullet_k {\delta^2_{\bullet i,\C_k}}}/(n^\bullet_k + 1)$ and
  $\Delta_{l,i}^{\bullet +} = {n^\bullet_l {\delta^2_{\bullet i,\C_l}}}/(n^\bullet_l - 1)$,
  where $n^\bullet_k = \abs{\C_k}$ is the number of observations currently assigned to $\C_k$
  and $\delta^2_{\bullet i,\C_k} =\norm{\X_i - \hat{\m}_k}^2$ is the squared Euclidean distance
  between $\bX_i$ and the center of $\C_k$. Our proposal modifies the computation of
  $\Delta_{k,i}^{\bullet -}$ and $\Delta_{l,i}^{\bullet +}$ to correspond to changes in $\mW_K$.
  We call these modified quantities $\Delta_{k,i}^-$ and
  $\Delta_{l,i}^+$. Of particular note   is how $n^\bullet_k$ and
  $\delta_{\bullet i,\C_l}$ are adapted.  
  Our modification for $n^\bullet_k$ changes to the 
  number of available observations $n_{kj}$ in $\C_k$ in  each feature.
  We define the modified squared distance between
  $\bX_i$ and $\hat{\bmu}_k$ as $\delta_{i,\C_k}^2 =
  \sum_{j=1}^pY_{ij}Y^{(k)}_j(X_{ij} - \hat{\mu}_{kj})^2 = \sum_{j=1}^p\delta_{ij,\C_k}^2$, where
  $Y^{(k)}_j = I(n_{kj} > 0)$ and $\delta_{ij,\C_k}^2 = Y_{ij}Y^{(k)}_j(X_{ij} -
  \hat{\mu}_{kj})^2$. We now state and prove the forms of $\Delta_{k,i}^-$ and $\Delta_{l,i}^+$:
  \begin{result}\label{result3}
    The increase in $\mW_K$ upon transferring observation $\X_{i'}$ into $\C_l$ is
    $\Delta_{l,i'}^+ = \sum_{j=1}^p n_{lj} \delta_{i'j,\C_l}^2 /(n_{lj}+Y_{i'j})$. Also, the
    decrease in $\mW_k$ by moving observation $\X_{i'}$ out of $\C_k$  is $\Delta_{k,i'}^-
    = \sum_{j=1}^p n_{kj} \delta_{i'j,\C_k}^2 /(n_{kj}-Y_{i'j})$.
  \end{result}
  \begin{proof}
    First, consider the increase in $\mW_k$ as $\C_l$ grows to $\C_{l'} = 
    \{\C_l,\X_{i'}\}$. In this case, the mean of the $j$th coordinate
    of the $l$th group changes to $\hat{\mu}_{l'j} = (n_{lj} \hat{\mu}_{lj} + Y_{i'j}X_{i'j})/(n_{lj}+Y_{i'j})$. For
    brevity, we denote $I[\X_i \in \C_l]$ as $I_i^{\C_l}$. Then
    \begin{equation*}
      \begin{split}
        \Delta_{l,i}^+ & = \sum_{i=1}^n I_i^{\C_{l'}} \delta^2_{i,\C_{l'}}  - \sum_{i=1}^n
        I_i^{\C_{l}} \delta^2_{i,\C_l}\\
        & =  \sum_{i=1}^n \sum_{j=1}^p \left[ I_i^{\C_{l'}} \delta^2_{ij,\C_{l'}}  -
          I_i^{\C_{l}}\delta^2_{ij,\C_l} \right]\\
        & = \sum_{i=1}^n \sum_{j=1}^p \left[ Y_{ij}(X_{ij} - \hat{\mu}_{l'j})^2 I_i^{\C_{l'}}  -
          Y_{ij}(X_{ij} - \hat{\mu}_{lj})^2 I_i^{\C_{l}} \right] \\
        & =  \sum_{j=1}^p \sum_{i=1}^n (Y_{ij}X_{ij}^2I_i^{\C_{l'}} - Y_{ij}X_{ij}^2I_i^{\C_{l}}) \\ &\qquad -2  
        \sum_{j=1}^p\sum_{i=1}^n (Y_{ij}X_{ij}\hat{\mu}_{l'j}I_i^{\C_{l'}} -
        Y_{ij}X_{ij}\hat{\mu}_{lj}I_i^{\C_{l}}) \\
        &\qquad \qquad + \sum_{j=1}^p \sum_{i=1}^n(Y_{ij}{\hat{\mu}_{l'j}}^2I_i^{\C_{l'}} -
        Y_{ij}\hat{\mu}_{lj}^2 I_i^{\C_{l}})
      \end{split}
    \end{equation*}
    The first term equals $\sum_{j=1}^p Y_{i'j} X_{i'j}^2$. The inner summation in the second term
    is
    \begin{equation*}
      \begin{split} 
        & \sum_{i=1}^n (Y_{ij}X_{ij}\hat{\mu}_{l'j}I_i^{\C_{l'}} -
        Y_{ij}X_{ij}\hat{\mu}_{lj}I_i^{\C_{l}}) \\
        &= (n_{lj}+Y_{i'j}) {\hat{\mu}_{l'j}}^2 - n_{lj} \hat{\mu}_{lj}^2 \\ &= (n_{lj}\hat{\mu}_{lj}
        + Y_{i'j}X_{i'j})^2/(n_{lj}+Y_{i'j}) - n_{lj} \hat{\mu}_{lj}^2 \\
        & = [(n_{lj}\hat{\mu}_{lj} + Y_{i'j}X_{i'j})^2 - n_{lj}(n_{lj}+Y_{i'j})
        \hat{\mu}_{lj}^2]/(n_{lj}+Y_{i'j})\\
        & = [2n_{lj}\hat{\mu}_{lj}Y_{i'j}X_{i'j} + Y_{i'j}X^2_{i'j} -
        n_{lj}Y_{i'j}\hat{\mu}_{lj}^2]/(n_{lj}+Y_{i'j}) \\
      \end{split}
    \end{equation*} 
    so that the second term is $-2\sum_{j=1}^p[2n_{lj}\hat{\mu}_{lj}Y_{i'j}X_{i'j} +
    Y_{i'j}X^2_{i'j} - n_{lj}Y_{i'j}\hat{\mu}_{lj}^2]/(n_{lj}+Y_{i'j})$. Similarly, the third term
    is $\sum_{j=1}^p[2n_{lj}\hat{\mu}_{lj}Y_{i'j}X_{i'j} + Y_{i'j}X^2_{i'j} -
    n_{lj}Y_{i'j}\hat{\mu}_{lj}^2]/(n_{lj}+Y_{i'j})$. Combining all
    three terms yields
    \begin{equation*}
      \begin{split}
         \Delta_{l,i}^+ 
        & = \sum_{j=1}^p \left[Y_{i'j} X_{i'j}^2 -
          \frac{2n_{lj}\hat{\mu}_{lj}Y_{i'j}X_{i'j} + Y_{i'j}X^2_{i'j} - n_{lj}Y_{i'j}\hat{\mu}_{lj}^2}
          {(n_{lj}+Y_{i'j})}\right]\\
        & = \sum_{j=1}^p \frac{n_{lj}Y_{i'j} X_{i'j}^2 - 2n_{lj}\hat{\mu}_{lj}Y_{i'j}X_{i'j}
          +n_{lj}Y_{i'j}\hat{\mu}_{lj}^2}{(n_{lj}+Y_{i'j})}\\
        & = \sum_{j=1}^p \frac{n_{lj}}{n_{lj}+ Y_{i'j}} Y_{i'j}(X_{i'j} - \hat{\mu}_{lj})^2\\
        & \equiv \sum_{j=1}^p \frac{n_{lj}}{n_{lj}+ Y_{i'j}} Y_{i'j}Y_j^{(l)}(X_{i'j} -
        \hat{\mu}_{lj})^2 =  \sum_{j=1}^p\frac{ n_{lj}}{n_{lj}+Y_{i'j}} \delta_{i'j,\C_l}^2.
      \end{split}
    \end{equation*}
    Similar calculations show the reduction in $\mW_K$ to be $\Delta_{k,i'}^-= \sum_{j=1}^p n_{kj}
    \delta_{i'j,\C_k}^2 /(n_{kj}-Y_{i'j})$.
  \end{proof}
  Our calculations provide the wherewithal for computing the changes
  in $\mW_K$  with incomplete records. We now
  detail the specific steps of our algorithm that flow from 
  \citet{hartigan1979algorithm} but uses the  
  derivations of Result~\ref{result3}. 
  \begin{enumerate}
  \item[Step 1:] \textbf{Initial assignments:} Obtain initializing values $\{
    \hat{\m}_{k}^{(-1)} ; k=1,2,\ldots,K \}$ using methods to be introduced in 
    Section~\ref{init}. Use these initial values to obtain $\bxi^{(0)} = (\xi_1^{(0)},
    \ldots,\xi_n^{(0)} )$ and $\bpsi^{(0)}=( \psi_1^{(0)}, \psi_2^{(0)}, \ldots, \psi_n^{(0)})$
    where $\xi_i^{(0)}=\argmin_{1 \le k \le K} \delta_{i,\C_k}^2 \text{   and   }
    \psi_i^{(0)}=\argmin_{1 \le k \le K;k \ne \xi_i^{(0)}} \delta_{i,\C_k}^2$  are the indices of
    the closest and second closest cluster means to $\X_i$. In general, let $\bxi^{(t)}$ denote
    the cluster assignment of every observation at iteration $t$. Let $\mbC^{(0)}$ be the
    partition defined by $\bxi^{(0)}$. Update $\hat{\m}^{(0)}$ given $\mbC^{(0)}$. 

  \item[Step 2:] \textbf{Live set initialization:} Put all cluster indices in the live set
    $\L$. Thus, $\{1,2,\ldots,K\} \in \L$.
  \item[Step 3:] \textbf{Optimal-transfer stage:} At the $t$th iteration, we have $\bxi^{(t)}$,
    $\bpsi^{(t)}$, and cluster means $\hat{\m}^{(t)}$. For each $i=1,2,...,n$, suppose that
    $\xi_i^{(t)}=k$. Next, do (a) or (b) according to whether $k$ is
    in the live set $\L$ or not:

    \begin{enumerate}
    \item {\bf Case ${(k \in \L)}$:}
      Let $k^* = \argmin_{b \ne k} \Delta^+_{b,i}$. If
      $\Delta^+_{k^*,i} \ge  \Delta^-_{k,i}$, leave $\X_i$ as currently assigned, setting
      $\xi_i^{(t+1)} = \xi_i^{(t)}$, leaving $\hat{\m}_k^{(t+1)}$ unchanged, and setting
      $\psi_i^{(t+1)} = k^*$. Otherwise transfer $\X_i$ to cluster $k^*$, setting $\xi_i^{(t+1)} =
      k^*$ and updating both $\hat{\m}_k^{(t+1)}$ and $\hat{\m}_{k^*}^{(t+1)}$. Also, assign
      $\psi_i^{(t+1)}=k$ and move cluster indices   $k$ and $k^*$ to the live set $\L$. 
    \item {\bf Case $\bm{(k \notin \L})$:} Do as in Step 3(a), but compute $\text{argmin}_{b \in
        \L} \Delta^+_{b,i}$, the minimum increase in $\mW_K$ only over the members of the live set. 
    \end{enumerate}

  \item[Step 4:] \textbf{Termination check:} The algorithm terminates
    if $\L \equiv \emptyset$, that is, $\L$ is empty. This happens if
    no transfers were made in Step     3. Otherwise,     proceed to Step $5$. 

  \item[Step 5:] \textbf{Quick transfer stage:} For each $i=1,2,\ldots,n$, let
    $\xi_i^{(t)} = k$ and $\psi_i^{(t)}=k^*$. We need not check observation $\bX_i$ if both $k$
    and $k^*$ have not changed in the last $n$ steps. If $\Delta^+_{k^*,i} \ge \Delta_{k,i}^-$, no
    change is necessary and $\xi_i^{(t+1)}$, $\psi_i^{(t+1)}$, $\hat{\m}_k^{(t+1)}$, and
    $\hat{\m}_{k^*}^{(t+1)}$ are left unchanged. Otherwise, we set $\xi_i^{(t+1)} = k^*$ and
    $\psi_i^{(t+1)}=k$, and update $\hat{\m}_{k}^{(t+1)}$ and $\hat{\m}_{k^*}^{(t+1)}$.

  \item[Step 6:] \textbf{Live set updates:}  Any cluster that is modified by
    the previous quick transfer step is added to the live set until at least the end of the next
    optimal-transfer stage. Any cluster not updated in the previous $n$ optimal-transfer steps is
    removed from the live set.

  \item[Step 7:] \textbf{Transfer switch:} If no transfer has taken place in the last $n$ 
    quick-transfer steps, return to Step 3 (Optimal-transfer). Otherwise, return to Step 5
    (Quick-transfer).
  \end{enumerate}

Our algorithm is a needed adaptation of
  \citet{hartigan1979algorithm} that accounts for the use of the 
  partial distance~\citep{himmelspachandconrad10} and $\mW_K$
  which,  Results \ref{result1} and   \ref{result2}, is the appropriate function to optimize. Our $k_m$-means algorithm prevents missing
  values from affecting estimation of the group means or contributing
  to the value $\mW_K$ for a given partition, but allows the observed
  features in incomplete records to be considered and
  grouped. Further, our approach  directly finds the locally best
  partition   minimizing $\mW_K$, while $k$-POD uses 
  a majorization function that is minimized at each iteration
  using a traditional $k$-means algorithm. We now provide some strategies for initialization.

  \subsection{Initialization}
  \label{init}
  Appropriate  $k$-means initialization can speed up convergence and 
  yield groupings with objective function close to the global minimum
  \citep{maitra09,jain2010data}. Although many initialization
  methods~\citep{astr70,milligan80,bradleyandfayyad98,maitra09,ostrovskyetal13} exist, 
  $k$-means++ is popular and relatively inexpensive and produces
  clusterings that are at worst $O (\log k)$ competitive with the optimal grouping
  \citep{arthur2007k}. In effect, $k$-means++ prefers initial centers
  that are appropriately spread out and has the following steps:
  \begin{enumerate}
  \item Set the first center, $\hat{\m}_1 = \X_i$, where $\X_i$ is chosen randomly from $\{\X_1,\X_2,\ldots,\X_n\}$.
  \item Initialize the $k$th ($k=2,\ldots,K$) group by choosing $\X_i$ with probability $p_i =
    {d_i^2}/({\sum_{i=1}^n d_i^2})$, where 
    $d_i^2=\min_{k=1,...,k-1}d_{i,k}^2$ and $d_{i,k}$ is some distance measure between $\bX_i$ and the $k$th cluster center $\hat{\m}_k$. 
  \item Repeat Step 2 until all $K$ centers have been initialized.
  \end{enumerate}
  For applying $k$-means++ with incomplete records, it would seem natural to 
  set $d_{i,k}^2$ to $\delta_{i,\C_k}^2$  as
  defined in Section~\ref{hw}. It turns out, however, that a more effective strategy is to
  use $\tilde d_{i,k}^2 = \tilde \delta_{i,\C_k}^2  = {\delta_{i,\C_k}^2}/{\sum_{j=1}^p
    Y_{ij}Y^{(k)}_{j}}$. The use of $\tilde d_{i,k}^2$ is related to
  adopting a \emph{partial distance    strategy} as in
  \citet{hathawayandbezdek01}. Note that $\delta_{i,\C_k}^2$ is not a
  true   distance measure, as the triangle inequality does not hold, and further $\delta_{i,\C_k }=0$
  implies only that $\X_i$ and $\hat{\m}_k$ are equal in the dimensions where both have recorded
  or calculated values, respectively. Denoting $\bY^{(k)} = (Y_1^{(k)},
  Y_2^{(k)},\ldots,Y_p^{(k)})$, and since $\E(\X_i\mid\X_i\in\C_k)=\m_k$ as in the development
  of Section~\ref{prelim}, we have  
  \begin{equation*}
    \begin{split}
      \E ( \delta_{i,\C_k}^2 & \mid \Y_i, \Y_k,\bX_i\in\C_k ) \\ 
      & = \E \left[ \sum_{j=1}^p Y_{ij}Y_{kj}
        (X_{ij} -  {\mu}_{kj})^2 | \Y_i, \Y_k, \bX_i\in \C_k  \right] \\  
      & = \sigma^2 \sum_{j=1}^p Y_{ij}Y_{kj}.
    \end{split}
  \end{equation*}
  Thus $ \tilde \delta_{i,\C_k}^2 $ provides a more appropriate measure of $\X_i$'s potential
  contribution to the error variance than $
  \delta_{i,\C_k}^2$. Figure~S-1
  provides execution times and clustering accuracy for
  selected simulation settings upon using  $ \delta_{i,\C_k}^2$ and  $
  \tilde \delta_{i,\C_k}^2$ and are discussed further in Section~\ref{res}. 
  In brief, they show 
shorter execution times and, on average, slightly more accurate
partitions with  $ \tilde \delta_{i,\C_k}^2 $-weighting. So we only use initializations with
  $\tilde{\delta}_{i,\C_k}^2$-weighting. For each clustering, we generate $100Kp$ initializations of our
  algorithm to account for the potential increase in the number of local
  minima with $K$ and $p$.   

  \subsection{Estimating the number of groups}

  In practice, $K$ is rarely known {\em a priori} and needs to be assessed from the dataset. There are
  many available methods \citep{krzanowskiandlai88, milligan1985examination,
    hamerly2003learning, pelleg2000x, maitraetal12} for estimating $K$ in the context of
  completely observed data. A  computationally inexpensive method that has performed well
  in many $k$-means contexts is the jump statistic of
  \citet{sugar2003finding}, so we adapt this approach for our setting. The development of the jump
  statistic is motivated by rate distortion 
  theory, with the number of groups estimated based on the rate of decrease of the
  average Mahalanobis distance between each observation and its assigned cluster center as $K$
  increases. In the usual $k$-means setting, the jump statistic chooses $\hat  K = \argmax_{ K
    \in \mK } \hat {D}_{\bullet, K}^{-p/2} -  \hat{D}_{\bullet,K-1}^{-p/2}$ where $\mK$ is the set
  of all values of $K$ under consideration, and the estimated distortions $\hat{D}_{\bullet,K} =
  \text{WSS}_K/np$ with  $\hat{D}_{\bullet,0}\equiv 0$. We have observed that merely replacing
  WSS$_K$ above with the optimized $\mW_K$ does not yield satisfactory results. Instead, we also
  replace $p$ in the average distortion and jump statistic calculations with the average
  effective dimension $\bar p$. Note that, as per Result~\ref{result2}, $\mW_K/n\bar{p}$ is
  a biased estimator of $\sigma^2$ given the true cluster assignments, and the MLE of $\sigma^2$
  under the assumptions of Result~\ref{result1}. Thus, our proposal to select the optimal
  $K$ chooses  $\hat  K =\argmax_{K\in\mK} J_k = \argmax_{K\in\mK} \hat D_K^{-\bar p/2} -
  \hat D_{K-1}^{-\bar p/2}$, with estimated distortions modified to be $\hat D_K = \mW_K/n
  \bar{p}$. As before, we set $\hat D_0 \equiv 0$. The use of a measure of effective dimension was
  initially suggested in \citet{sugar2003finding} for cases with strong dependence between
  features. Simulations indicate that using $\bar{p}$ in place of $p$ for missing data yields an
  improved estimator for $K$ and also improves partitioning performance. 
\section{Performance Assessments} \label{sim}
We first illustrate and evaluate our methodology on the classification
dataset introduced in  \citet{wagstaff2004clustering}. We next perform a
comprehensive simulation study to evaluate the different aspects of
our algorithm. Performance in all cases was measured  numerically and
displayed graphically. Our methods and its competitors were evaluated
in terms of the  Adjusted Rand index ($AR$)
\citep{hubert1985comparing}. The $AR$ index is commonly used as a 
measure of agreement between two clusterings, in this case between the
true cluster labels and the labels returned by either clustering
method. The index attains a maximum value of 1 if the two partitions
are identical and has an expected value of zero when the partitioning
has been done by chance. 

\subsection{Illustration on SDSS Data}\label{kiri}
The Sloan Digital Sky Survey (SDSS) contains  millions of observations 
of astronomical bodies, but the subset used in
\citet{wagstaff2004clustering} -- and that we use to evaluate
performance -- has observations from $1220$ galaxies and $287$
stars. The five included features are 
brightness (measured in psfCounts), size (in
petroRad, with some negative values for reasons that are not
entirely clear), a measure of
texture, and two measures of shape (M{\_}e1 and M{\_}e2), which we will
refer to as {\em Shape1} and {\em Shape2} in our analysis. The dataset
is complete but for $42$ galaxies that are missing both measures of
shape. 
Figure S-2
displays the dataset, with color corresponding to the true classifications of star or galaxy. 
Many of the features are heavily skewed, while the shape measures are
predominantly 
marked by very long tails both in the left and right directions. 

The $k_m$-means algorithm chooses homogeneous groups with spherical
spreads, so we first transform, center, 
and scale each feature. For brightness and 
texture we use a $\log$ (base $10$) transformation but the other 
features contain negative values, so, for these variables, we use
the inverse hyperbolic sine transformation \citep{burbidgeetal88}
given by $h(u;\theta) = \arcsinh (\theta u)/\theta$ for $\theta \neq 0$
and $h(u;0)= u$. For the three variables, we chose $\theta=10$ which
substantially moderates the skewness and peakedness. 
(Other trial values of $\theta$  indicated insensitivity
of our results to small changes.) The transformed data
were then centered and scaled by the 
sample standard deviation of each feature. 
For the  SDSS data, $k$-POD and $k_m$-means yield identical
clusterings. Because this is a classification dataset, classifications
are available. 
So we  examine the clustering returned by $k_m$-means for $K=2$ 
groups for its ability to distinguish between
stars and galaxies in the entire data set, ability to distinguish
between stars and galaxies in the incomplete observations, and the
effect of deleting incomplete observations. 

The $K=2$ $k_m$-means (and $k$-POD) clustering has an $AR$ index of $0.988$. 
\begin{figure}[h]
\centering
\includegraphics[width=0.65\textwidth]{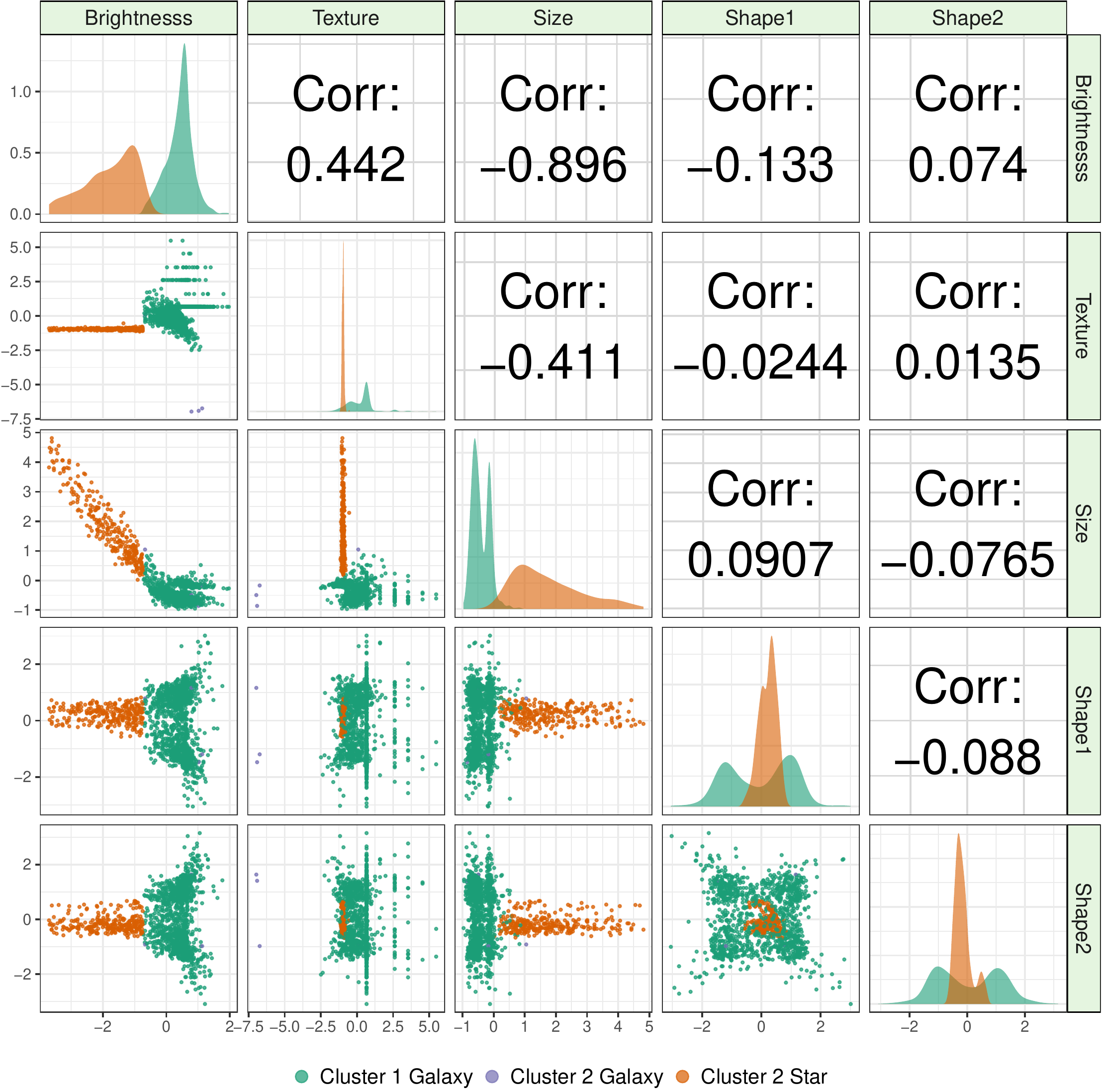}
\caption{Scatterplots, univariate densities, and correlations of the transformed
 features of the SDSS data. The three colors correspond to the observed combinations 
 of the two clusters specified by the $K=2$ $k_m$-means clustering and
 the true classification of each observation. Note that there are only
four misclassifications, of galaxies (identified in blue).}
\label{fig:sdss}
\end{figure}
Figure \ref{fig:sdss} displays the results of clustering
in the transformed variable space are shown in the scatterplot matrix of Figure \ref{fig:sdss}. (Here, color indicates
membership in the final grouping from $k_m$-means, with shading
corresponding to the observation's true classification.)
\begin{table}[h]
  \centering
  \caption{Confusion matrix of the $k_m$-means (also $k$-POD)
    groups versus the true classes in the SDSS dataset.} 
  \label{tab:SDSS}
  \begin{tabular}{lrr}\hline
    Group & Stars & Galaxies\\ \hline
    1     & 287   &   0 \\
    2 & 4 & 1216\\
    \hline
  \end{tabular}
\end{table}
Further, Table~\ref{tab:SDSS} provides a {\em  confusion matrix} or
cross-tabulation of the frequency of stars and galaxies classified in
each $k_m$-means (or $k$-POD) group. 
If one considers the clusters as classifying stars and
galaxies, only $4$ galaxies are 
misclassified. Further, every galaxy (and therefore every
observation) with missing values
is correctly classified. In this case, $k$-means clustering on only
the complete observations results in an identical partition when
comparing the grouping of only the complete observations but using
$k_m$-means on the entire datasets. Thus, for these data, 
$k_m$-means is able to correctly cluster the incomplete observations,
but that the inclusion of the incomplete observations has no effect on
the clustering of the complete observations. Given that the incomplete
observations make up a small percentage of the larger of the two
clusters, this lack of a difference in class assignments is not
unexpected.

The performance of $k_m$-means (and $k$-POD) is much better than that of any
of the methods 
reported in Figure 1a of \citet{wagstaff2005making}, where the best
performer for $K=2$ had $AR \approx 0.2$. While we are unable to identify
the reasons for the poorer performance reported in that paper, it is
probable that our transformation to remove skewness and subsequent
scaling may have had an important role in our better performance. 
Despite the transformations, it is clear that the features
are not independent. There is clear separation between the two
clusters in both brightness and size, which are strongly negatively
correlated.  Thus other approaches to clustering may also be
worth pursuing for this dataset. However, this dataset offers a
valuable illustration of $k_m$-means and indicates promise. We now proceed to evaluate the performance of $k_m$-means in several large-scale simulation
experiments. 

\subsection{Simulation Studies}
\subsubsection{Experimental Framework}\label{framework}
We thoroughly evaluated the $k_m$-means algorithm on a series of
experiments encompassing different missingness mechanisms, clustering
complexities,  dimensions,  numbers of clusters,
 proportions of missing values, and  missingness 
 mechanisms. We discuss these issues next.
 \paragraph{\bf Clustering Complexity of Simulated Datasets}
 The $k_m$-means algorithm inherently assumes data
from homogeneously and spherically dispersed groups, so we restricted our
attention to this framework. Within this setting, we 
simulated grouped data of different clustering complexities as measured by the
generalized overlap measure and 
implemented in the {\tt C} package {\sc CARP} 
\citep{melnykovandmaitra11} or the {\tt R} package {\sc MixSim}
\citep{melnykov2012mixsim}, which is what we used in this paper. 
The generalized overlap (denoted by $\mathring\omega$ here) is a single-value summary of the
pairwise overlap \citep{maitraandmelnykov10}  between $K$ groups and is
defined as $(\omega^{(1)}-1)/(K-1)$, where $\omega^{(1)}$ is the
largest eigenvalue of the matrix of pairwise overlaps between the
groups. $\mathring\omega$  takes higher values for greater overlap
between groups ({\em i.e.}, when there  is higher 
clustering complexity) and lower values when the groups are
well-separated. We simulated clustered data of different
dimensions ($p$), different numbers of groups ($K$), different sample
sizes ($n$) and different proportions ($\lambda$) of incomplete
records. For each $(K, n, p, \lambda, \mathring\omega)$,
we considered four different mechanisms of missingness which led to
the incompleteness of the records, as discussed next.

\paragraph{\bf Missingness Mechanisms} 
Missing data are traditionally categorized into one of three different
types: missing 
completely at random (MCAR), missing at random (MAR), and not missing at random (NMAR)
\citep{little2014statistical}. For MCAR data, the probability
that an observation record is missing a feature measurement depends
neither on the 
observed values nor on the value of the missing data point. To simulate 
MCAR data, we randomly removed data independent of dimension and
cluster. On the other hand, under MAR, the
probability that an observation has a missing coordinate may depend on the
observed values, but not on the true values of the missing data point. To
simulate data under the MAR mechanism, we followed \citet{chi2016k} in
randomly removing data in only $40\%$ of the dimensions, leaving the
other features completely observed. Note that in order to obtain a
specific overall proportion $\lambda$ of missing data,
the proportion of data missing in the partially observed dimensions
will be higher -- in our case $2.5 \lambda$. When 
the probability that a value is missing depends on its true value, we
say that the data are NMAR. We considered two different mechanisms for
NMAR, namely NMAR1 for data that are 
MCAR but only in specified clusters, and NMAR2 that follows
\citet{chi2016k} in missing the bottom quantiles of each dimension in
specified clusters. As with the MAR scenario, a higher proportion
of data than $\lambda$ had to be removed in the partially
observed groups to obtain the desired overall proportion of missing
data.

Our experimental setup thus had a multi-parameter setup, with
values as per  Table~\ref{table:simpars}.
\begin{table}[h]
\caption{Values for each parameter used in simulation study.}
\label{table:simpars}
\centering
\begin{tabular}{l | l}
\hline
Parameter & Values \\
\hline
\# groups $(K)$ & $4,7$ \\
\# observations $(n)$ & $500,1000,5000$ \\
Dimension $(p)$ & $5,10$ \\
Missing proportion $(\lambda)$ & $0.05,0.1,0.2,0.3$ \\
Overlap $( \mathring{\omega} )$& $0.001,0.01,0.05$ \\
Missingness mechanism & MCAR, MAR, NMAR1, NMAR2 \\
\hline
\end{tabular}
\end{table}
The {\sc CARP} and {\sc MixSim} packages afford the possibility of
providing general assessments of clustering performance in different
settings. Therefore, for each of $(K, n, p, \lambda,
\mathring\omega)$ and missingness mechanism, we generated 50 synthetic 
datasets within the given experimental paradigm in order to
assess performance. Thus, our simulation consisted of a total of
28,800 simulated datasets. 

\paragraph{\bf Additional  details regarding implementation}
We first compared performance of $k_m$-means with $k$-POD (through its {\tt
  R} package {\sc kpodcluster}) because both $k_m$-means and
$k$-POD are geared towards optimizing \eqref{wss} so can provide a
direct comparison. We compared performance of the two algorithms for
the case when $K$ is known in terms of execution speed, as well
as (and more importantly) clustering efficacy measured in
terms of $AR$. $k$-POD is naturally slower than $k_m$-means, due to
its repeated application of the $k$-means algorithm at each iteration.  
Therefore, we used $100Kp$ initializations for $k_m$-means but only
$5$ initializations for $k$-POD. The performance gains for $k_m$-means
from using  $100Kp$ initializations rather than $5$ initializations of
$k_m$-means 
are in most cases minimal, and the use of  unequal numbers of
initializations across methods reflects how each would most likely
be used in practice (with $k$-POD being used with a number of
initializations that make it practical to apply.) We
also evaluated performance of our modified jump 
statistic in deciding $K$.We chose our
candidate $K$s to be in the set $\{1,2,\ldots,2K_\bullet \}$ where
$K_\bullet$ was the true $K$ under which the particular simulated dataset was
obtained. We restricted our
use of the jump statistic estimator to
$k_m$-means because the slower performance of $k$-POD makes
application of $k$-POD for each candidate $K$ to be too
computationally onerous to evaluate. We now report and discuss
performance. 

\subsubsection{Results}\label{res}
We first discuss performance of $k_m$-means and $k$-POD when $K$ is
known and follow with clustering performance with $k_m$-means for when
$K$ is unknown and estimated using our modified jump statistic.

\paragraph{\bf Performance with Known $K$}
We evaluate performance of $k_m$-means and $k$-POD in terms of
execution time, initialization and clustering performance. 

{\bf Execution Times:}
\begin{figure}
\centering
\includegraphics[width=.75\textwidth]{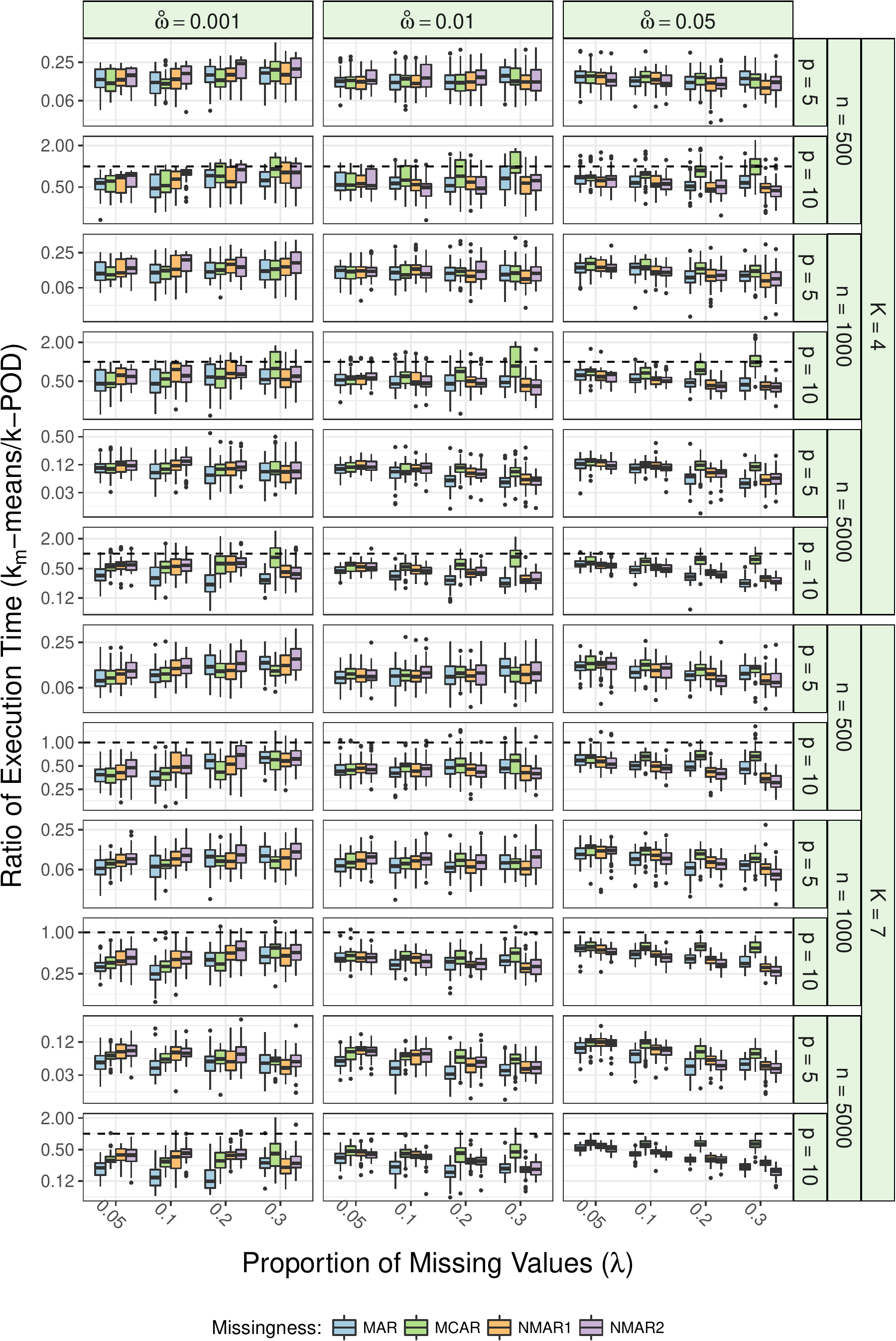}
\caption{Ratio of execution times between $k_m$-means with $100Kp$
  initializations each and $k$-POD  with 5 initializations each for
  50 simulated data sets in each simulation setting. 
  The semi-solid line at 1 on the vertical axis is
  the point above and 
  below which $k_m$-means takes more and less time, respectively,
  than $k$-POD.}
\label{fig:time}
\end{figure}
Figure~\ref{fig:time} displays the ratio of the execution times of $k$-POD
and $k_m$-means, with known $K$. Even though
$k_m$-means uses a far greater number of 
initializations (on the scale of $20kp$), it is in nearly all cases at least as fast as $k$-POD. Paired $t$-tests of the
execution times show that $k_m$-means is significantly faster at an $\alpha=0.05$ level in all
but two settings. Much of this difference can be attributed to the efficiency of the
algorithms. In essence, for each initialization, $k$-POD must perform an entire $k$-means
routine at each iteration, whereas $k_m$-means handles missing data within one $k$-means
routine. Thus, with the exception of some special cases discussed in
Section~\ref{comparisons}, one would expect $k_m$-means to be
more efficient. (Note, however, that in these few cases, $k_m$-means
would still have been faster if both $k_m$-means and $k$-POD were to
have been run with the same number of
initializations. Indeed, Figure~S-3 
which reports the
per-initialization run relative gain of $k_m$-means over $k$-POD also
supports this conclusion.)

{\bf Initialization:}
As described in Section~\ref{init}, Figures S-1a and S-1b
respectively display the execution times and performance of our 
$k_m$-means algorithm when initialized
 using $ \delta_{i,\C_k}^2$- and  $ \tilde 
\delta_{i,\C_k}^2$-weighting for 
selected settings. In general, initialization done using  $\tilde
\delta_{i,\C_k}^2$-weighting leads to faster clustering than that when using  
$\delta_{i,\C_k}^2$-weighting. In terms of final clustering performance, $\tilde
\delta_{i,\C_k}^2$-weighting leads to better results than
$\delta_{i,\C_k}^2$-weighting. 
This improved speed and performance is most
pronounced in the case of lower $K$. In the few settings where
$\delta_{i,\C_k}^2$-weighting is marginally better on average, the
$AR$s for those clusterings are more widely-dispersed. It 
is interesting to note that of the two NMAR methods, clusterings on NMAR2 data are 
more accurate within each weighting relative to NMAR1. In this paper, we only report
results from $k_m$-means using the $\tilde\delta_{i,\C_k}^2$-weighting.

\begin{figure} 
  \centering
\includegraphics[width=0.75\textwidth]{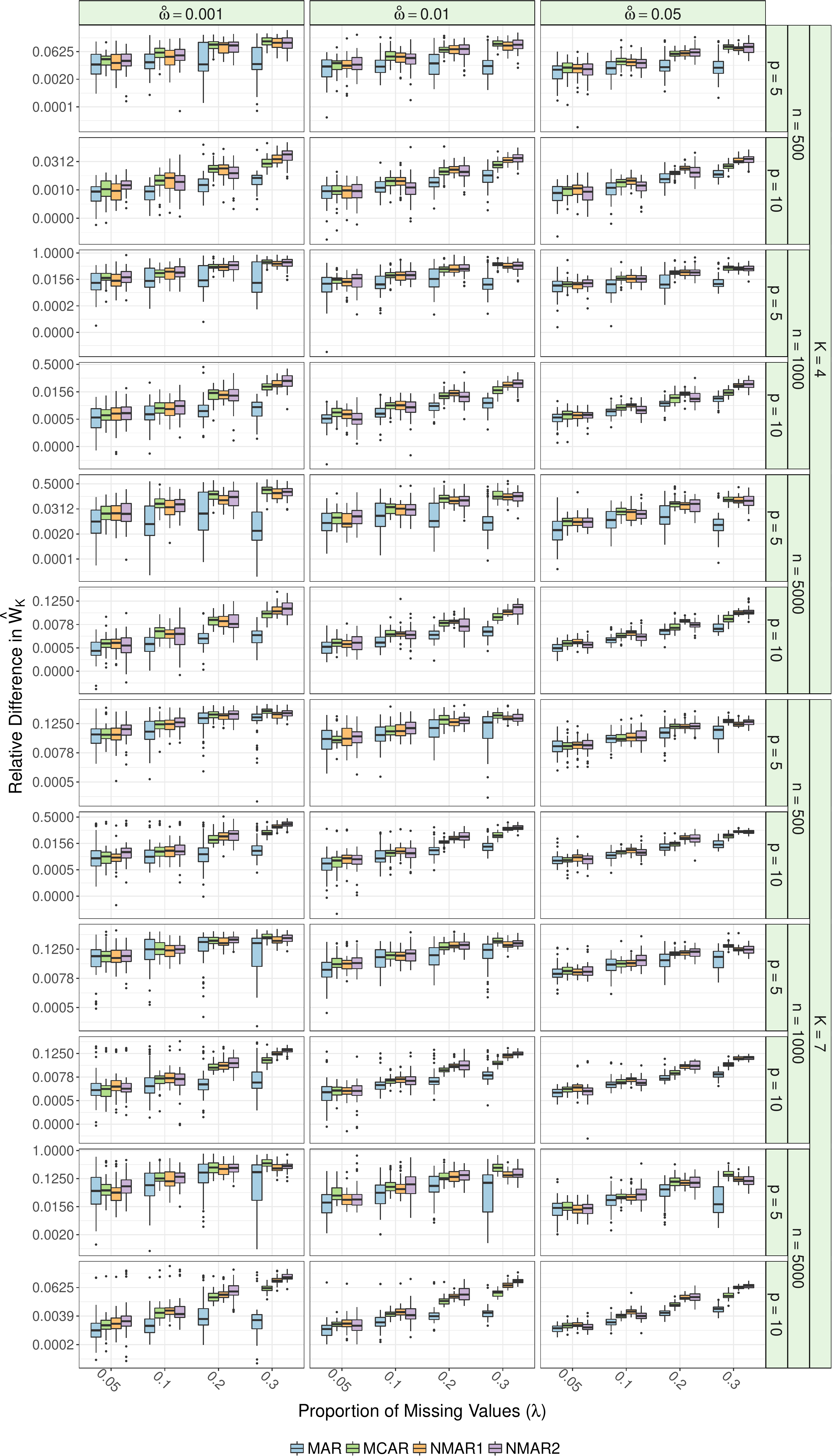}
\caption{Relative difference in optimized objective function $\hat 
  \mW_K$ obtained using $k$-POD and $k_m$-means for 50 simulated data sets in each simulation setting.}
\label{fig:rel-simwss}
\end{figure}

{\bf Overall Comparisons:}
\label{comparisons}
Figure~\ref{fig:rel-simwss} provides the relative decrease in the
optimized $\hat \mW_K$ upon using $k_m$-means over  the $k$-POD algorithms. 
In general, the optimization improves more with
deviations from MCAR as well as with increasing proportions of missing
observations. This, despite $k_m$-means' execution times that are a
fraction of the $k$-POD execution times (Figure~\ref{fig:time}). Thus,
in terms of optimizing ~\eqref{wss}, $k_m$-means is uniformly a better
\begin{figure*}
  \centering
\includegraphics[width=0.8\textwidth]{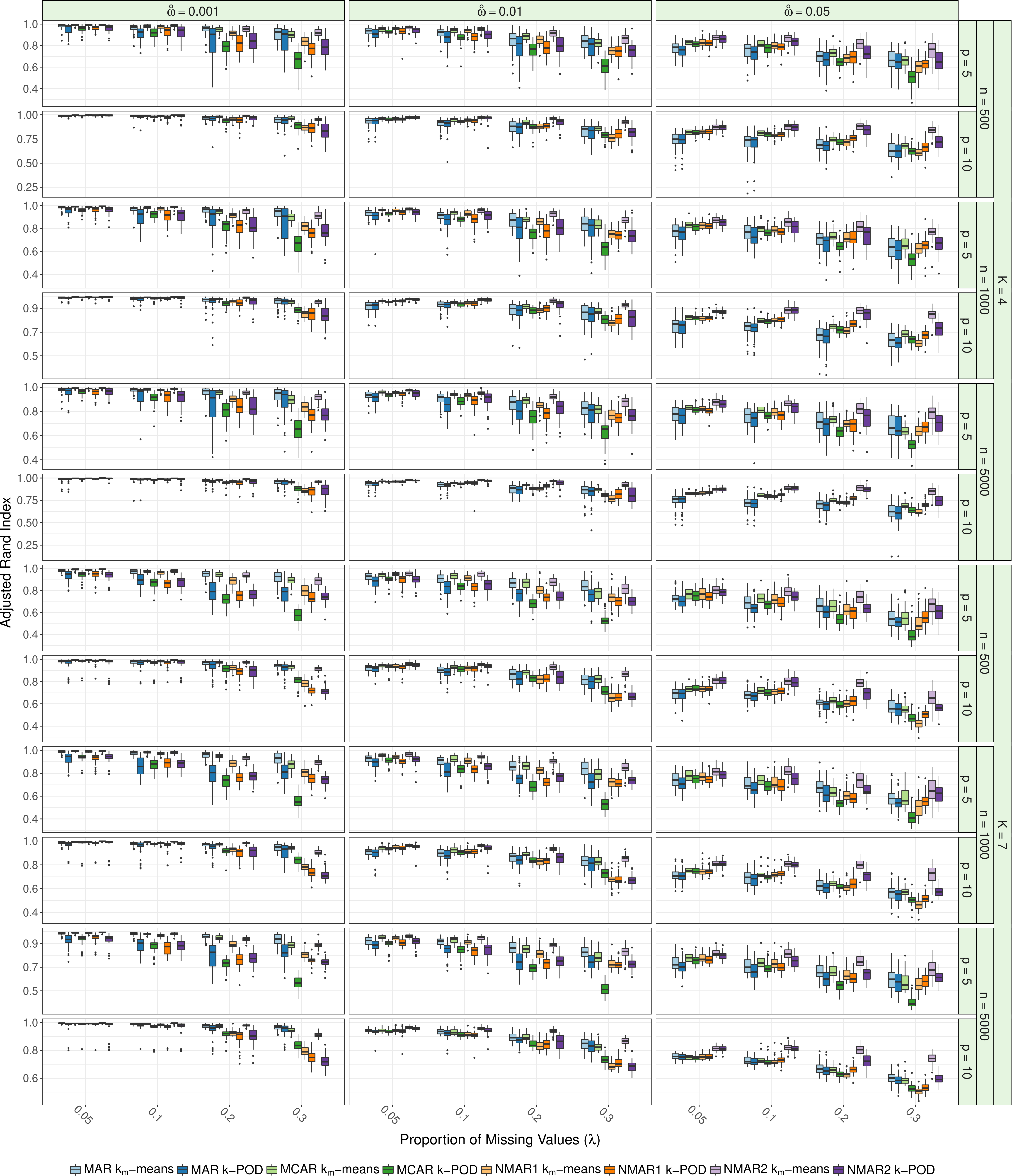}
      \caption{Adjusted Rand Index (AR) for $k_m$-means and $k$-POD for 50 simulated data sets in each simulation setting.}
\label{fig:simar}
\end{figure*}
performer than $k$-POD. Figure~\ref{fig:simar} summarizes clustering performance in terms of
$AR$ for each setting. Excluding NMAR1 data, $k_m$-means performs at
least as well as $k$-POD, and significantly better as per one-sided
paired t-tests in all but 5 settings. However, $k$-POD is the better 
performer in the NMAR1 cases with moderate to high clustering
complexity (overlap) and with larger proportions of missing
observations. We hypothesize that that is  because the NMAR1 mechanism
with large proportions of incomplete records and higher overlap
results in unbalanced designs 
and estimated non-spherical clusters: in such a scenario, optimizing
\eqref{wss} may not be synonymous with finding the best
clustering. This hypothesis is supported by Figure~\ref{fig:rel-simwss}
that indicates that the terminal (local minimum) value obtained by $k_m$-means is
lower than that obtained by $k$-POD.  
Further, the difference in execution times between the
two methods for MCAR data shrinks as the proportion of incomplete records increases. As expected,
performance for both methods suffers with increasing proportion of missing data
or clustering complexity, but both methods perform admirably even outside of
MCAR data. Finally, we note that the higher number of initializations
used for $k_m$-means can not explain its superior performance over 
$k$-POD: Figure~S-4 
shows very similar results when only 5 initializations were used for both methods.
\paragraph{\bf Estimating $K$ via the modified jump statistic}
Because of the slower performance of $k$-POD, we only evaluate
performance using $k_m$-means in this section. 
Figure~\ref{fig:khat} shows that the jump statistic often correctly estimates
$K$, but underestimates it in more difficult cases, particularly with larger
$K$. Errors
are most strongly correlated with the proportion of missing data and cluster overlap, with
poor estimation of $K$ when $K$, $\mathring{\omega}$, and $\lambda$ are at their highest
values. There is noticeable difficulty in estimating $K$ in MAR data
at $K=7$, but in each case, results improve with increasing
$n$. (Recall that MAR data is missing significant proportions of 
values in selected features.)
\begin{figure}[h]
  \centering
  \includegraphics[width=0.75\textwidth]{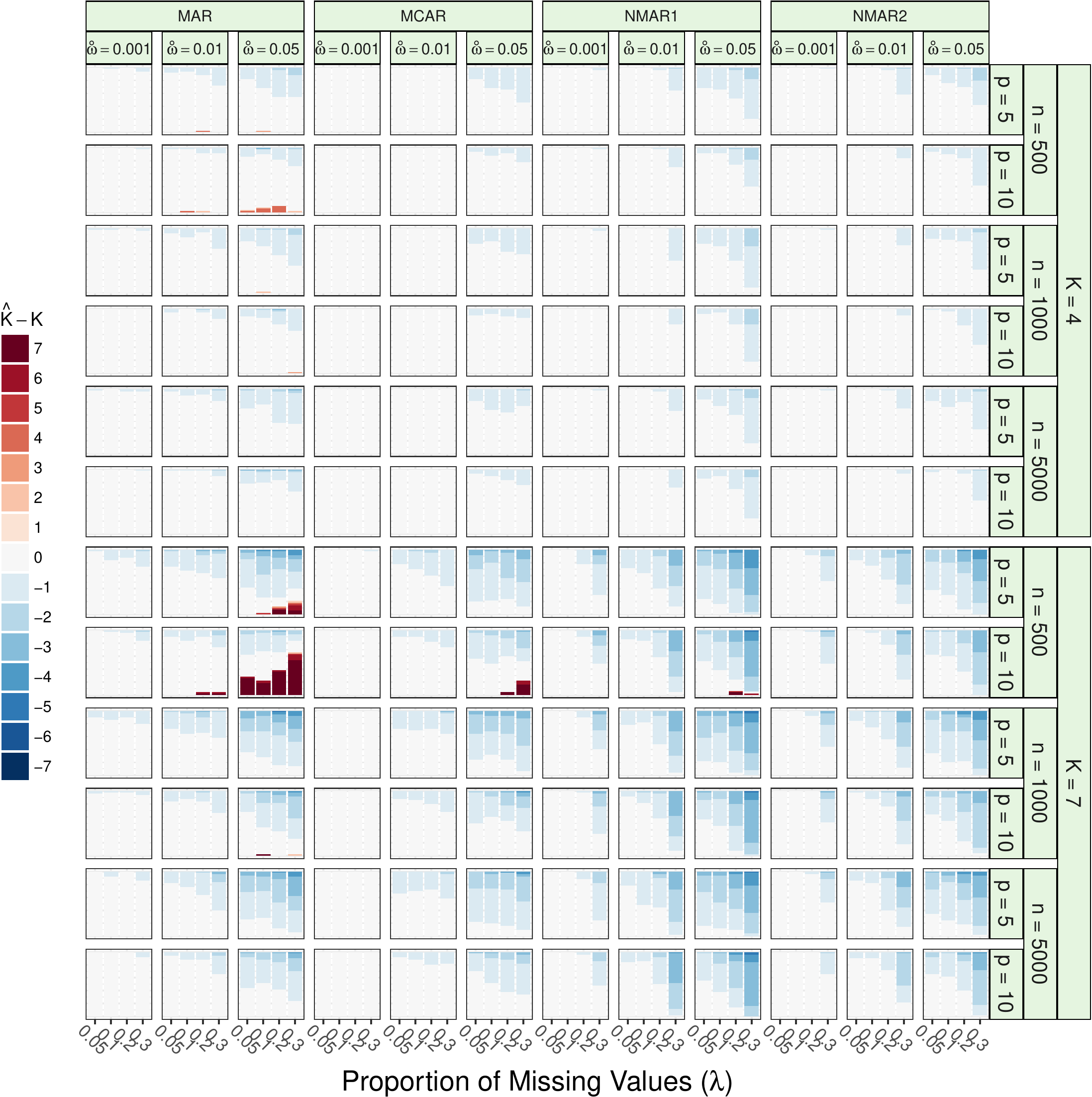}
  \caption{ Difference between estimated number of clusters ($\hat{K}$) by the modified jump statistic 
  and the true number of simulated clusters ($K$) for 50 simulated data sets in each simulation setting.}
\label{fig:khat}
  \end{figure}
\begin{figure}
  \centering
  \includegraphics[width=0.85\textwidth]{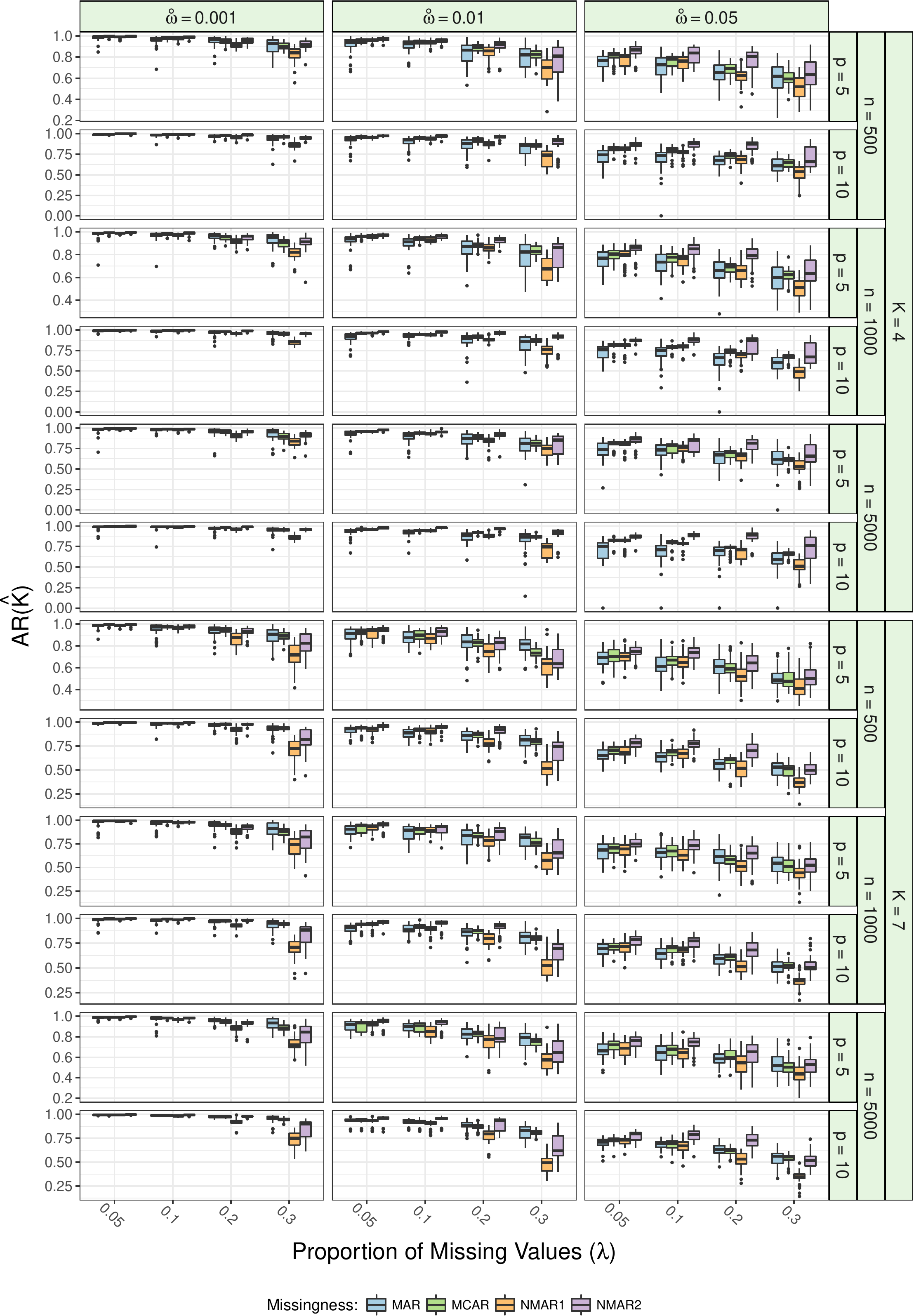} 
  \caption{Adjusted Rand Indices ($AR(\hat{K})$) from $k_m$-means
    clusterings using the Jump-estimated number of clusters  $\hat K$ 
    for 50 simulated data sets in each simulation setting.}
\label{fig:khatar}
\end{figure}
 When data are heavily missing in this
exact manner, but the 
partially observed features are known to be of importance, it may be more appropriate to use
the soft constraints approach of \citet{wagstaff2004clustering}.  We
also observe a tendency to 
underestimate $K$ in each NMAR setting. This is to be expected,
possibly even desired, because the 
NMAR settings may end up removing the majority of the values in
clusters selected to be partially 
observed. Particularly as the overlap between clusters increases, it
is not surprising that 
the jump estimator would underestimate $K$, and instead assign to
nearby clusters the remaining observations with 
high proportions of missing values. Thus, we see
limited improvement in 
$\hat{K}$ as $n$ increases in the NMAR settings.
Figure~\ref{fig:khatar} plots the
$AR$ of the final clusterings using $\hat{K}$ and confirms
our observations drawn from 
Figures~\ref{fig:simar} and \ref{fig:khat}. The observed $AR$s using
$\hat{K}$ tend to be less than or equal to the $AR$ obtained using
$K_\bullet$ but the differences are not large, with $AR\approx 1$ for
lower values of $\mathring{\omega}$ and $\lambda$. We also see that 
in many cases, the $AR$ value for NMAR1 data is lower than those from other types of
missingness. This can be traced back to the tendency to underestimate $K$ in NMAR1 data in
particular.

\subsubsection{Comparison with Imputation Methods}
\begin{figure*}
  \centering
  \includegraphics[width=\textwidth]{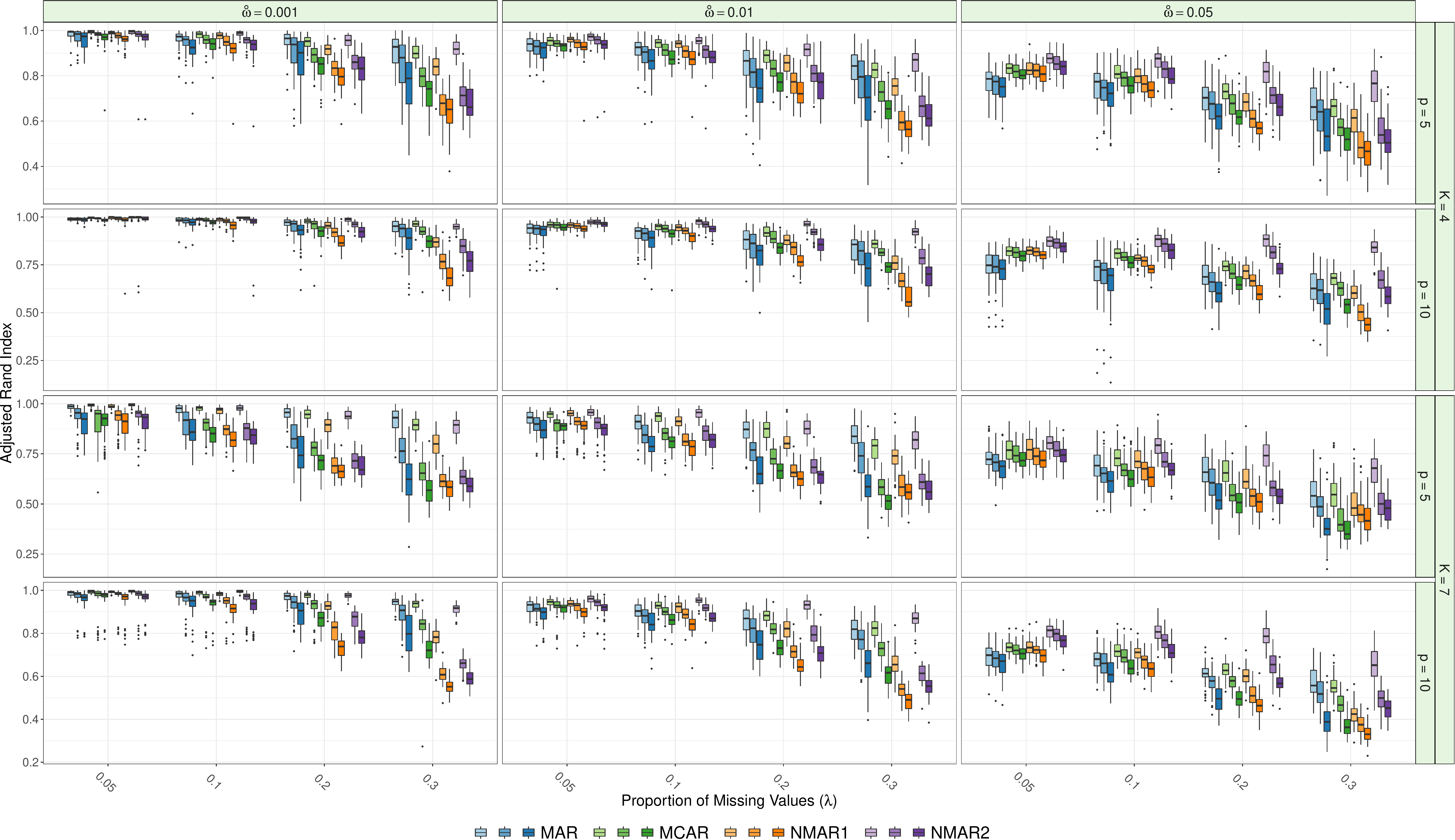}
\caption{Performance of imputation methods and $k_m$-means on
  different simulation settings with known   $K$ and for $n=500$. For
  each missingness mechanism in the figure, the light shades depict
  $k_m$-means while the medium and dark shades are for Amelia- and 
  mice-imputed $k$-means groupings.}
\label{fig:ar-imp}
\end{figure*}
We have also, at the helpful suggestion of the Editor, provided a
somewhat more limited comparison with  two imputation methods. Our
comparisons were only limited in that they were done for datasets with
$n=500$ simulated records and known $K$ but having all the other simulation settings
of Section~\ref{framework}. We compared $k_m$-means with two popular
imputation methods {\em Amelia} \citep{honaker2011amelia}
and and {\em mice} \citep{buuren2011mice}, both of which are
implemented in R packages having the same respective names. 
Both  methods draw imputations of the missing values. {\em Amelia}
uses the Expectation-Maximization algorithm~\citep{dempsteretal77} on
multiple bootstrapped samples of the original incomplete data to
obtain estimates of the complete-data parameters, and then uses these
bootstrapped estimates to replace the missing  values with imputed
values. On the other hand, {\em mice} draws imputations of the
missing data points using chained equations. 

For both imputation methods, we used the default settings to
obtain imputed values for the missing observations and then used
$k$-means on the results. 
Figure~\ref{fig:ar-imp} shows the clustering performance of both
imputation methods relative to $k_m$-means.
We see a consistent improvement in accuracy of clustering using
km-means with the largest difference in AR observed when the
censoring methods stray further from assumptions made in {\em Amelia}
and {\em mice},
for instance in the NMAR2 case, and also as the proportion of missing
values increases. This deficiency of the imputation methods is not
surprising because neither method directly uses the inherent grouped
structure of the data and only provides imputations based on an
overall (even though sophisticated) 
view of the data. Clustering algorithms are then used on the filled-in
data values. 

Our large-scale simulation experiments show that  our $k_m$-means
algorithm performs well over several different cluster sizes,
missingness mechanisms, and proportions of missing values.  Our
modified jump statistic  is also effective in selecting 
the number of groups. We now apply our methods to identify the kinds
of activation in a finger-tapping experiment. 
\section{Identifying kinds of activation in a finger-tapping experiment} \label{fmri}
\label{application}
\begin{table*}
\centering
\caption{Number of voxels recording values in each replication in the fMRI experiment.}
\label{table:fmridata}
\begin{tabular}{l  c  c c c c c c c c c c }
\hline 
replication & 1 & 2 & 3 & 4 & 5 & 6 & 7 & 8 & 9 & 10 & 11\\
\hline
\# voxels & 2103 & 2413 & 2666 & 1731 &  2378 & 1543 &  2583 &  2408 &  1834 & 894 & 1251 \\
\hline
\end{tabular}
\end{table*}

Functional magnetic resonance imaging~(fMRI) is a noninvasive tool
used to determine 
cerebral regions that are activated in response to a particular stimulus or task
\citep{bandettinietal93, belliveauetal91, kwongetal92, ogawaetal90}. The simplest experimental
protocol involves acquiring images while a subject is performing a task or responding to a
particular stimulus, and relating the time course sequence of images
(after correction and pre-processing) to the expected response to the
input stimulus \citep{fristonetal94, lazar08}. 
However, there are concerns about the inherent reliability and reproducibility of the
identified activation \citep{maitraetal02, gullapallietal05, maitra09a}. \citet{maitra10}
illustrates an example of differing activation maps obtained over twelve different
sessions, where the same  subject performed a simple finger-tapping
task in each session.  
We seek to combine activation maps across each experiment to help understand the nature of brain activation
in this experiment. It would be advantageous to have the ability to incorporate results from different 
fMRI studies without the need to re-analyze each experiment. Next, we show that this problem can be cast as an
incomplete-records clustering problem.

Our data set for this experiment is from a left-hand finger-tapping experiment of a right-hand-dominant 
male and was acquired over twelve regularly-spaced sessions over the course of
two months. Each data set was preprocessed and voxel-wise $Z$-scores were obtained that
quantified the test statistic under the hypothesis of no activation at
each voxel.  
The $Z$-scores from each session
were thresholded using cluster-thresholding methods
\citep{wooetal14}. Because of 
concerns that the normally-right-hand-dominant male subject may have
been inadvertently tapping his right hand fingers 
\citep{maitra10}, the activation statistics for one session were
dropped from our study. 
Thus, there are a total of eleven replicated test statistics. Our
interest is then in 
classifying the voxels using their corresponding activation test statistics. Note that because of the
thresholding, activation statistics are not available across all replicates.
Table~\ref{table:fmridata} lists the number of voxels above thresholding at each replication.
There are 2827 total voxels that were identified as activated in at least one session, with a
maximum of five missing values across replications. There are only $156$ voxels without any
missing values. Thus, our goal is to cluster voxels based on the $Z$-scores of eleven
replications, where incomplete records arise because, after
thresholding, not all replications have a $Z$-score for each voxel. 
The use of $Z$-scores and assumption of independence over
replications because of the substantial time between any two 
sessions makes this an ideal case for the assumption of homogeneous
spherical dispersions for each sub-population of voxels. 

\begin{figure}[h]
\mbox{
\subfloat{\includegraphics[width=0.48\textwidth]{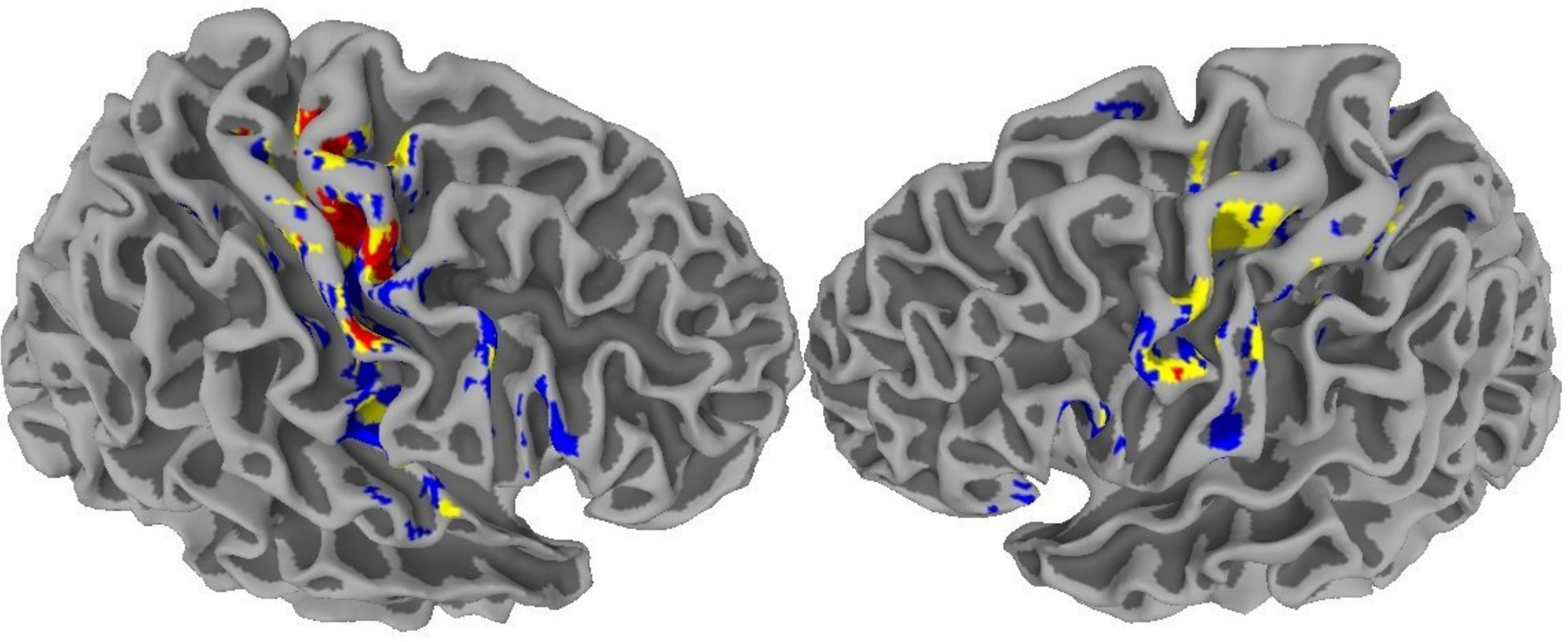}}}
\mbox{
\subfloat{\includegraphics[width=0.48\textwidth]{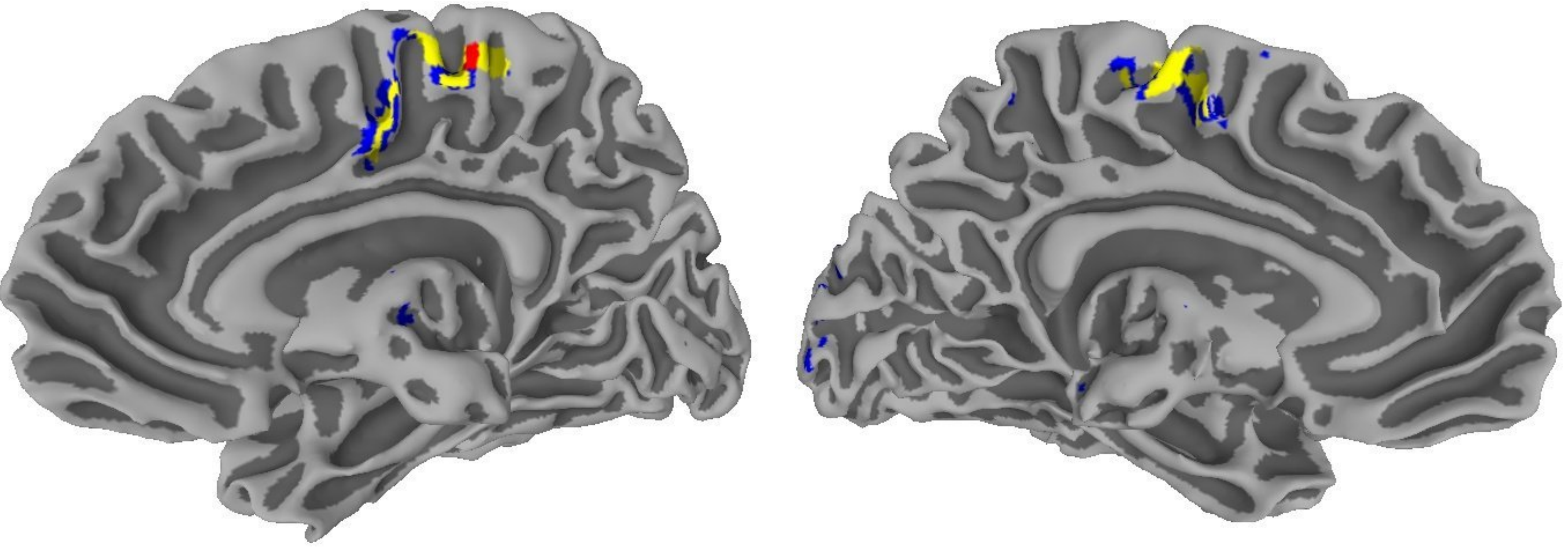}}}
\caption{Regions of activation detected from voxels in eleven
  replications of an fMRI study.}
\label{fig:fmri}
\end{figure}

As before, we
run the $k_m$-means algorithm to termination, with  
$100p\max(K,10)$ initializations using the methods in Section~\ref{init},
for $K=1,2,\ldots,20$. The jump statistic identified the three-groups
solution as the optimal partitioning. 
The resulting groups are displayed in Figure~\ref{fig:fmri} separately, for each of the eleven
experiments. The first group (denoted by red) consists of 235 voxels whose average mean 
$Z$-score is 10.31, the second (yellow) group has 965  voxels with average mean $Z$-score
6.83, and the third (blue) group includes 1627 voxels with an average mean $Z$-score of 4.96.
The first group is where the activation is most emphatic and is almost entirely in the right 
primary motor cortex (M1), the ipsi- and contra-lateral pre-motor cortices (pre-M1), and the 
supplementary motor cortex (SMA). The other two groups of voxels represent two different kinds
of milder activation and are primarily located in the right pre-SMA, and interestingly also in
the left M1, pre-M1, and the SMA. This last observation is an interesting finding and is
suggestive that activation in a right-hand dominant male is also associated in the left
hemisphere of the brain even when it is the non-dominant hand that is
active in performing a task. It is important to
note that following a whole data strategy in this experiment would not have been able to
identify this additional finding because almost all the 156 voxels that have non-thresholded 
$Z$-scores for all eleven replications (that is, having no missing
values) are in the right hemisphere. Our application here also
demonstrates an important approach to amalgamating the results from
different fMRI activation studies.

\section{Discussion}
\label{discussion}
We have extended the Hartigan-Wong $k$-means clustering algorithm to
the case for datasets that have incomplete records.  We do so by by
defining a (partial) distance measure and objective function that ignores
missing feature values. The modified objective
function necessitates adapting \citet{hartigan1979algorithm}'s algorithm
to account for incomplete records. We call the resulting algorithm
$k_m$-means. We also provide 
modifications to the $k$-means++ initialization method and the jump
statistic for estimating  the  
number of clusters. {\sc C} code implementing our methods is available
at \href{https://github.com/maitra/km-means}{https://github.com/maitra/km-means}. Our development represents an intuitive
addition to the body of work seeking to avoid 
discarding partially observed data or imputing
data. Simulations show this is an efficient 
and effective method for handling missing values, and application to
astronomical data yielded 
results in line with expectations. The $k_m$-means approach was also valuable in the analysis of
fMRI data, where the vast majority of observations (voxels) were treated as partially observed,
and located in the same area. Our proposed methods in this paper thus
provide a practical approach to $k$-means-type clustering in the
presence of incomplete observations. 

There are a number of issues that might benefit from further
attention. In the context of $k_m$-means, there is a need for more
research into initialization schemes for partially observed data. In
addition to considering weighting schemes such as $\delta_{i,k}^2$ and
$\tilde{\delta}_{i,k}^2$, alternative methods 
for choosing the initial center, $\hat{\m}_1$, may also lead to improved results. For example, we
may limit $\hat{\m}_1$ to only completely observed data, or assign weights for choosing
$\hat{\m}_1$ proportional to how many observed values each $\X_i$
has. Early results indicate 
each of these strategies lead to comparable results in most cases.
The use of $k$-means and Euclidean distances for applications such as
in the case of clustering of
GRBs~\citep{chattopadhyay2017gaussian,chattopadhyayandmaitra18} is not always
appropriate. Therefore, appropriate adjustments are required for
handling non-spherically dispersed groups of data or datasets with
unequal variances. However, there is scope for optimism because, as
pointed out by J. Bezdek, \eqref{wss} used the Euclidean norm, but the
proofs would generally hold for any inner product norm leading to
possibly hyper-ellipsoidal-shaped groups. Additionally, the
derivations would be fairly tractable even for the $\ell_1$ and 
$\ell_\infty$ norms. Finally, we note that it may also be possible to extend
the general approach  
of this paper to data containing observations with repeated
measures. Thus, while we have provided an efficient
algorithm for finding homogeneous spherically-dispersed clusters in
the case of incomplete records, several issues requiring further
attention remain.


\ifCLASSOPTIONcaptionsoff
  \newpage
\fi

\bibliographystyle{IEEEtran}

\bibliography{debib,references}

\renewcommand\thefigure{S-\arabic{figure}}\setcounter{figure}{0}
\renewcommand\thetable{S-\arabic{table}}
\renewcommand\thesection{S-\arabic{section}}
\renewcommand\thesubsection{S-\arabic{section}.\arabic{subsection}}
\renewcommand\theequation{S-\arabic{equation}}
\newpage
\section*{Supplementary Materials}
\captionsetup{singlelinecheck=on}
\begin{figure*}[h]
  \centering
  \mbox{\subfloat[]{\includegraphics[width=0.926\textwidth]{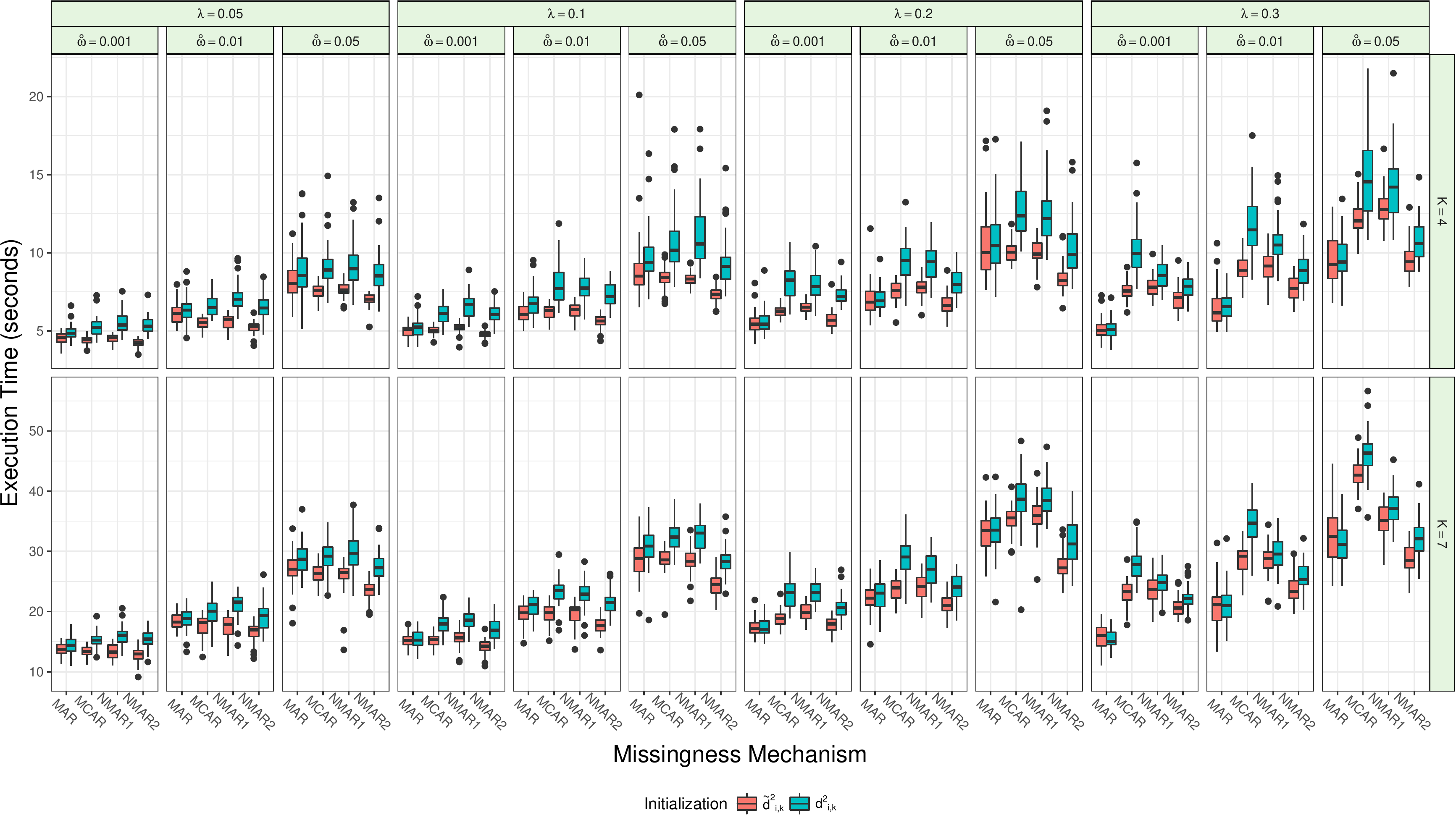}\label{fig:timecomp}}}
  \mbox{  \subfloat[]{\includegraphics[width=0.926\textwidth]{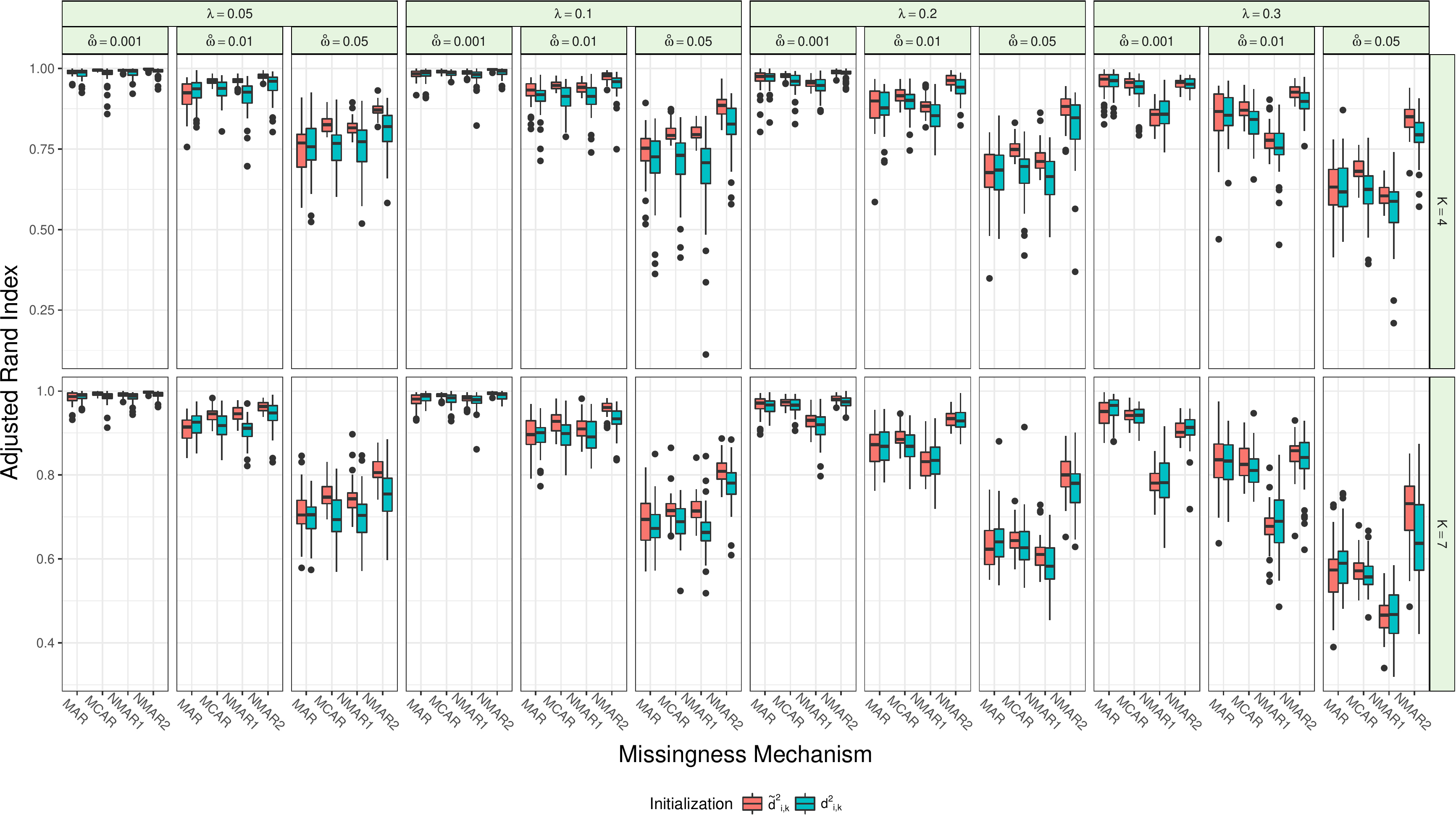}\label{fig:arcomp}}}
  \caption{(a) Execution times and (b) Adjusted Rand Indices of
    $k_m$-means using $ \delta_{i,\C_k}^2$- and  $ \tilde  
\delta_{i,\C_k}^2$-weighting for initializations. Only settings with $p=10$ and $n=1,000$ are shown.}
\label{fig:artimecomp}
\end{figure*}
\begin{figure}
\centering
\includegraphics[width=\textwidth]{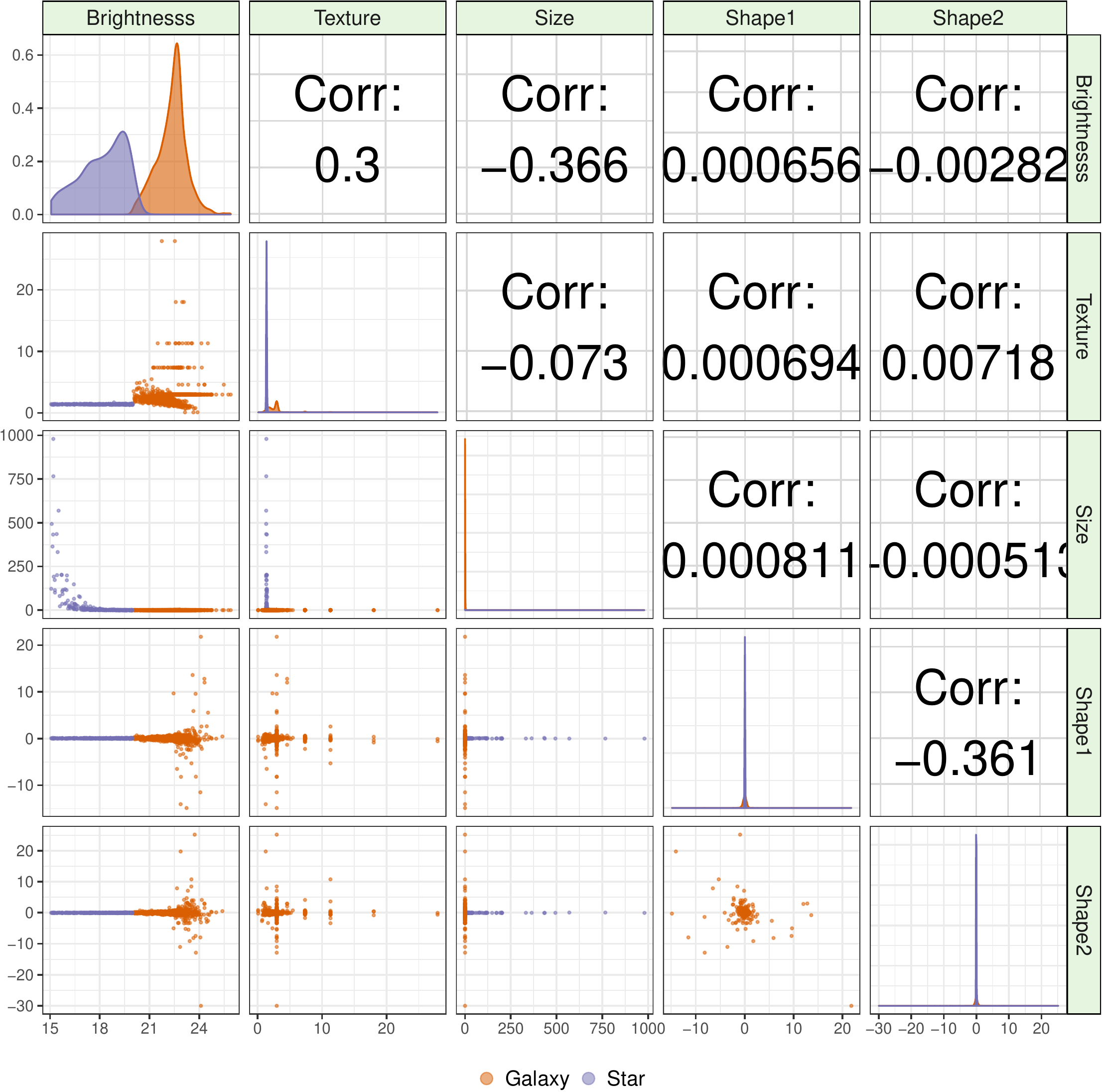}
\caption{Scatterplots, univariate densities, and correlations of the untransformed
 features of the SDSS data. 
The two colors correspond to the true classifications of each observation.}
\label{fig:sdss-untransformed}
\end{figure}

\begin{figure*}
\centering
\includegraphics[width=0.8\textwidth]{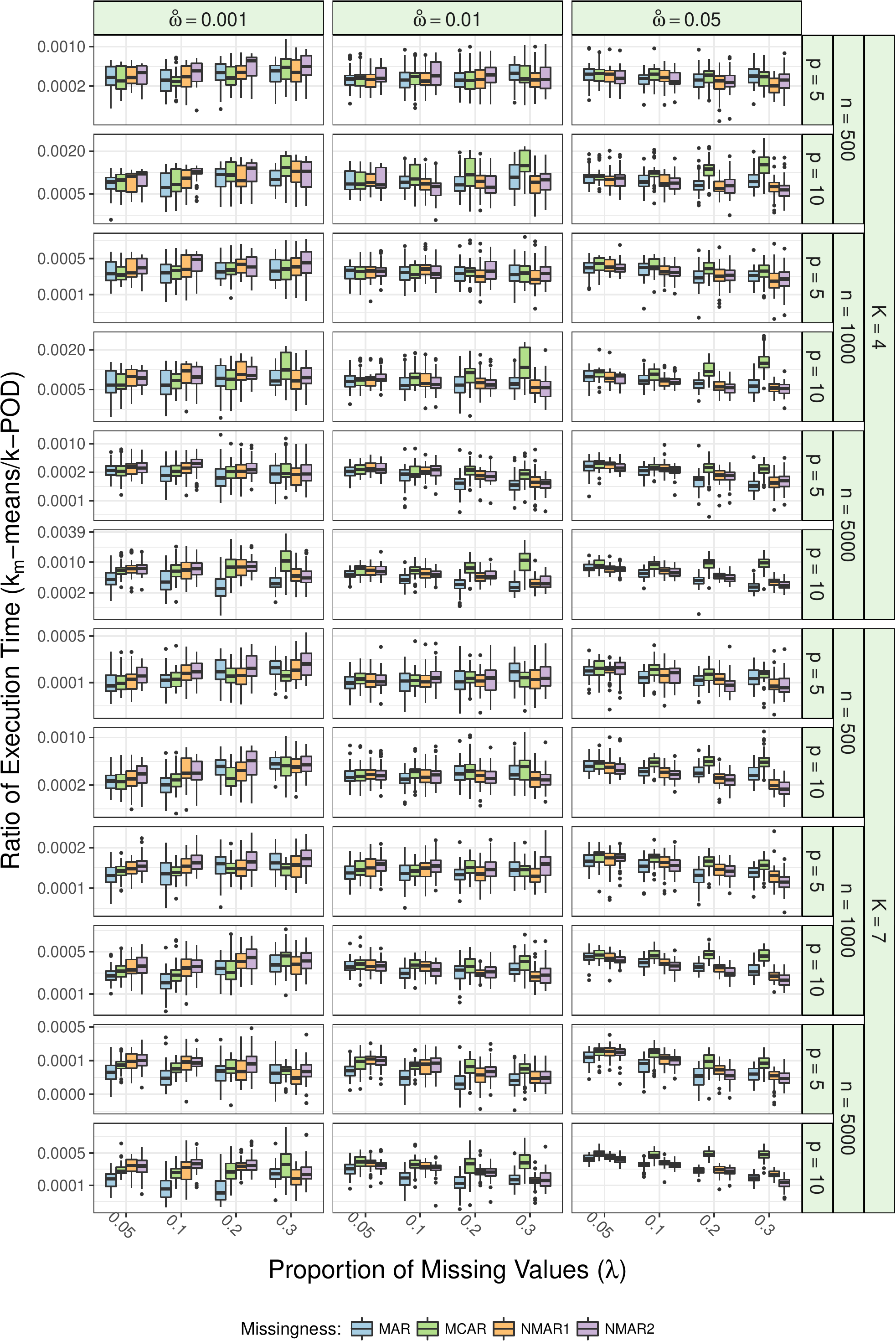}
\caption{Per-initialization execution time of $k_m$-means relative to
  $k$-POD.}
\label{fig:timeperinit}
\end{figure*}

\begin{figure*}
\centering
\includegraphics[width=\textwidth]{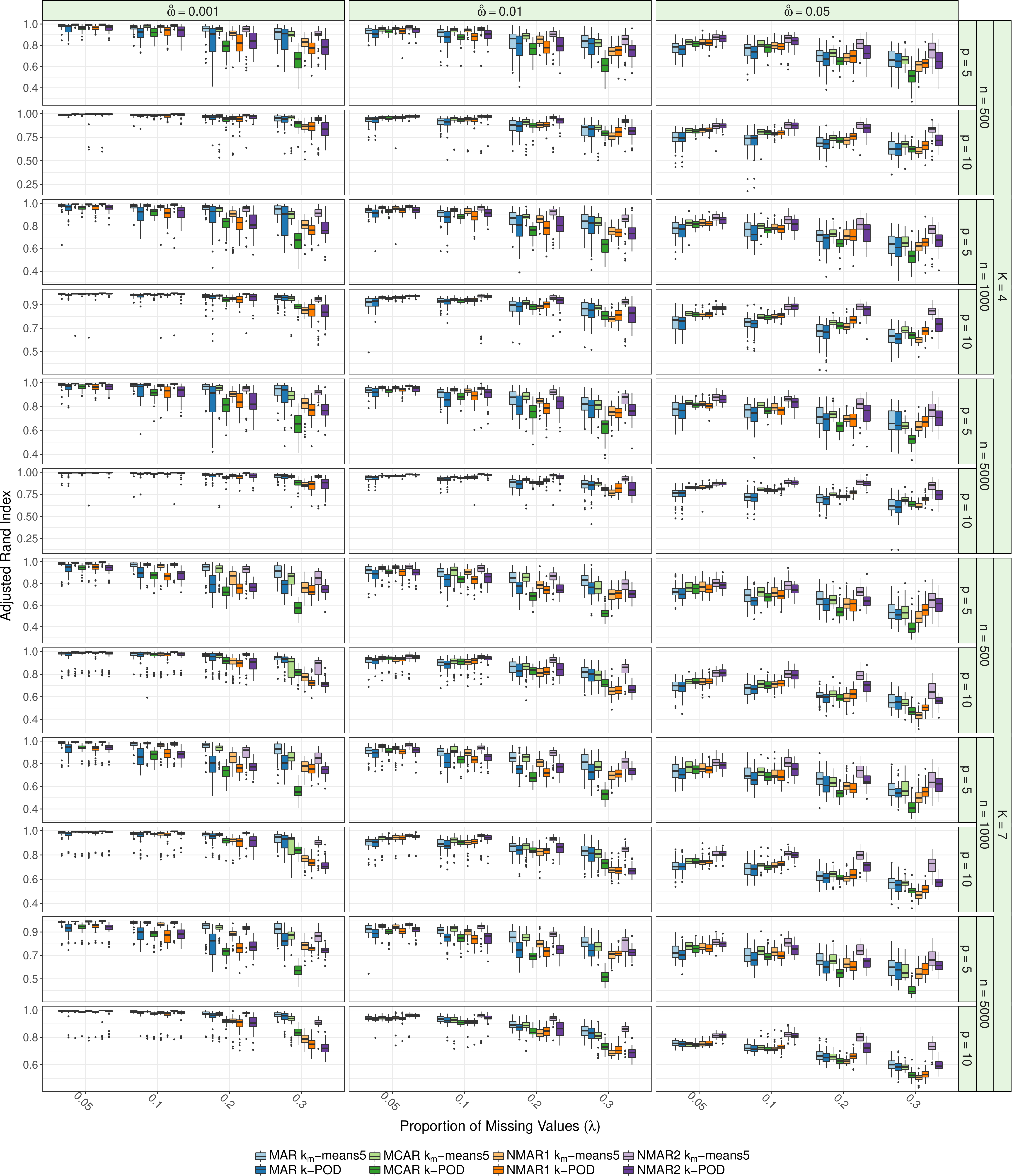}
\caption{Performance of $k_m$-means and $k$-POD when both methods were
  initialized 5 times.}
\label{fig:ar-km5}
\end{figure*}

\end{document}